\documentclass[10pt,twocolumn,letterpaper]{article}

\usepackage[pagenumbers]{cvpr} % To force page numbers, e.g. for an arXiv version

\definecolor{cvprblue}{rgb}{0.21,0.49,0.74}
\usepackage[pagebackref,breaklinks,colorlinks,allcolors=cvprblue]{hyperref}
\usepackage{amssymb}
\usepackage{graphicx}
\usepackage{cuted}      % provides the strip environment
\usepackage{float}
\usepackage{makecell}
\usepackage{caption}
\usepackage{subcaption}
\usepackage{pifont}
\usepackage{multirow}
\usepackage[utf8]{inputenc}
\usepackage{colortbl} % For table coloring
\usepackage{xcolor} % For text coloring
\usepackage{tcolorbox} % For block boxes
\usepackage[table]{xcolor}
\usepackage[dvipsnames]{xcolor}
\newcommand{\cmark}{\textcolor{cvprblue}{\ding{51}}}% ForestGreen
\newcommand{\xmark}{\textcolor{lightgray}{\ding{55}}}% red
\usepackage[symbol]{footmisc}

\def\paperID{} % *** Enter the Paper ID here
\def\confName{CVPR}
\def\confYear{2026}

\title{Embodied4C: Measuring What Matters for Embodied \\ Vision-Language Navigation}

\author{
Tin Stribor Sohn$^{1,4\dagger*}$ \hspace{8pt}Maximilian Dillitzer$^{2,4*}$ \hspace{8pt}Can Kandil$^{3,4*}$ \\ Jason J. Corso$^{5,6}$ \hspace{8pt}Eric Sax$^1$ \\
$^1$ Karlsruhe Institute of Technology \hspace{8pt}
$^2$ UAS Esslingen \hspace{8pt}
$^3$ TU Wien \\
$^4$ Dr. Ing. h.c. F. Porsche AG \hspace{8pt}
$^5$ University of Michigan \hspace{8pt}
$^6$ Voxel51 Inc. \\
{\tt\small tin\_stribor.sohn@porsche.de}
}

\begin{document}
\maketitle
\footnotetext[1]{Equal contribution; $^\dagger$Corresponding author.}
\begin{strip}
    \centering
    \includegraphics[width=\textwidth]{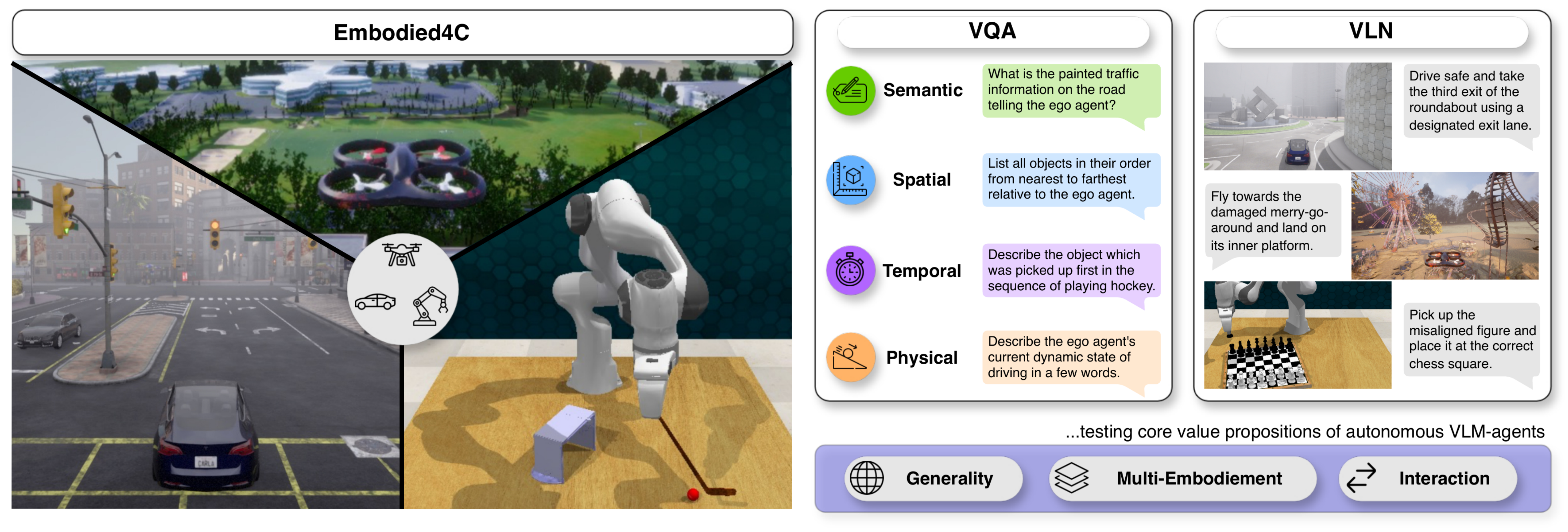}
    \captionof{figure}{\textbf{Overview of the Embodied4C benchmark.} Embodied4C spans three embodiment domains (autonomous driving, aerial navigation, robotic manipulation) and four embodied scenario understanding capabilities (semantic, spatial, temporal, physical). The benchmark evaluates autonomous agents in visual question answering and vision-language navigation for scenario understanding and control execution, and includes domain-far questions to probe generalization (i.e., sensor set, general knowledge, weather, scenario). This design targets core value propositions of vision-language models as embodied agents: generality, multi-embodiment competence, and natural interaction.}
    \label{fig:firstPage}
\end{strip}

%%%%%%%%%%%%%%%%%%%%%%%%%%%%%%%% ABSTRACT %%%%%%%%%%%%%%%%%%%%%%%%%%%%%%%% % DONE (MD)
\begin{abstract}
    Vision-language navigation requires agents to reason and act under constraints of embodiment. While vision-language models (VLMs) demonstrate strong generalization, current benchmarks provide limited understanding of how embodiment---i.e., the choice of physical platform, sensor configuration, and modality alignment---influences perception, reasoning, and control. 
    We introduce \textbf{Embodied4C}, a closed-loop benchmark designed as a Turing test for embodied reasoning. The benchmark evaluates the core embodied capabilities of VLMs across three heterogeneous embodiments---autonomous vehicles, aerial drones, and robotic manipulators---through approximately 1.1K one-shot reasoning questions and 58 goal-directed navigation tasks. These tasks jointly assess four foundational dimensions: semantic, spatial, temporal, and physical reasoning. 
    Each embodiment presents dynamic sensor configurations and environment variations to probe generalization beyond platform-specific adaptation. To prevent embodiment overfitting, Embodied4C integrates domain-far queries targeting abstract and cross-context reasoning. 
    Comprehensive evaluation across ten state-of-the-art VLMs and four embodied control baselines shows that cross-modal alignment and instruction tuning matter more than scale, while spatial and temporal reasoning remains the primary bottleneck for reliable embodied competence.
\end{abstract}

%%%%%%%%%%%%%%%%%%%%%%%%%%%%%%%% INTRO %%%%%%%%%%%%%%%%%%%%%%%%%%%%%%%% % DONE (MD)
\section{Introduction}
Vision-language models (VLMs) are increasingly applied in autonomous systems to unify perception, reasoning, and control through a shared language-based interface~\cite{Hwang_2024_arXiv_EMMA,Guo_2024_arXiv_DriveMLLM}. In vision-language navigation (VLN), this integration facilitates interpretable decision-making, open-ended instruction following, and context-aware behavior. Yet, existing evaluation practices rarely capture the core prerequisite for reliable performance in embodied environments: the capability for robust \emph{scenario understanding}.

Scenario understanding encompasses the ability to represent and reason about a situation across four core dimensions: \textit{semantic}, \textit{spatial}, \textit{temporal}, and \textit{physical} understanding~\cite{Sohn_2025_arXiv_Framework4C}. While existing benchmarks typically emphasize downstream task performance, they rarely isolate these foundational capabilities, making it difficult to attribute model behavior to specific reasoning deficits or strengths~\cite{Sohn_2025_CVPR_Drive4C}. Furthermore, many benchmarks are susceptible to \emph{inductive bias}: narrow task distributions, static sensor configurations, multiple-choice questioning, and fixed language templates allow models to overfit, yielding high scores without demonstrating true generalization~\cite{Chandak_2025_arXiv_Answermatching,Zheng_2024_arXiv_MCQBias}.

To address these limitations, we introduce \textbf{Embodied4C}, a closed-loop benchmark designed to evaluate the embodied reasoning capabilities of VLMs in visual question answering (VQA) and VLN. Conceptually, it serves as a \emph{Turing test for embodied intelligence}, assessing whether models can exhibit human-like understanding when perception, language, and action are integrated. In contrast to prior work that is restricted to only one domain~\cite{Sohn_2025_CVPR_Drive4C,Liu_2023_arXiv_AerialVLN,Gong_2023_arXiv_ARNOLD}, our benchmark spans multiple embodiments---autonomous ground vehicles, unmanned aerial vehicles (UAVs), and robotic manipulators---to systematically assess embodiment sensitivity and generalization. Embodied4C probes the four aforementioned reasoning capabilities through two complementary tasks: (1)~\emph{VQA} to assess scenario understanding, and (2)~\emph{VLN} to evaluate control and decision-making in realistic simulation environments. Both tasks are conducted in a closed-loop setting, ensuring that model outputs have causal impact on the environment.

A central feature of the benchmark is its additional focus on the \emph{value propositions of embodied agents}~\cite{Gemini_Robotics}. Specifically, the benchmark measures: (i) ~\emph{Generality}, operationalized through domain-far queries, embodiment changes, and domain shifts; (ii)~\emph{Embodiment-awareness}, captured via varying sensor modalities, placements, and physical configurations; and (iii)~\emph{Interactivity}, realized through closed-loop evaluation of VQA and VLN where perception, reasoning, and control are tightly coupled. To our knowledge, Embodied4C is the first benchmark to explicitly probe these propositions alongside the four core reasoning capabilities.

\noindent This work makes the following key contributions:
\begin{itemize}
    \item We introduce Embodied4C, the first closed-loop benchmark framework that unifies evaluation of core VLM value propositions across heterogeneous embodiments, while systematically probing semantic, spatial, temporal, and physical reasoning capabilities. 
    \item We develop a structured, exam-inspired, one-shot task design that operationalizes these evaluations through embodiment generalization, domain shifts, and domain-far reasoning challenges, mitigating overfitting and enabling granular attribution of embodied competencies.
    \item We provide a unified evaluation across three embodiment domains (ground vehicles, UAVs, robotic manipulators) to analyze the impact on perception, reasoning, and control on free-form VQA to reduce inductive answer biases.
    \item We conduct a comprehensive evaluation of ten pretrained VLMs~\cite{FastVLM_2025_Apple,Qwen2.5-VL,Google_2025_Gemma3,Meta_2024_Blog_Llama4,Anthropic_2024_Website_Claude3,Anthropic_2025_Claude4,OpenAI_2024_arXiv_GPT-4o,OpenAI_2025_GPT5} and four domain-specialized models~\cite{Jiang_2024_arXiv_Senna,Gao_2025_arXiv_OpenFly,Kim_2024_arXiv_OpenVLA,Lee_2025_arXiv_MolmoAct}, demonstrating both strengths and limitations in scenario understanding and embodied autonomous navigation.
\end{itemize}

\noindent Embodied4C establishes a novel evaluation framework for measuring what matters in embodied VLN, supporting the development of robust, general-purpose autonomous agents. All code will be open sourced upon publication.
%%%%%%%%%%%%%%%%%%%%%%%%%%%%%%%% RELATED WORK %%%%%%%%%%%%%%%%%%%%%%%%%%%%%%%% % DONE (MD)
\begin{table*}[t!]
    \centering
    \footnotesize
    \setlength{\tabcolsep}{8pt}
    %\resizebox{\textwidth}{!}{
    \begin{tabular}{r|cccc|cccc}
        \toprule
        \textbf{Benchmark} 
        & \textbf{SEM} & \textbf{SPA} & \textbf{TEM} & \textbf{PHY} & \textbf{GEN} & \textbf{EMB} & \textbf{INT} & \textbf{VLN} \\
        \midrule
        DriveMLLM~\cite{Guo_2024_arXiv_DriveMLLM}                       & \xmark & \cmark & \xmark & \xmark & \xmark & \xmark & \cmark & \xmark \\
        SpatialRGPT~\cite{Cheng_2024_NeurIPS_SpatialRGPT}               & \xmark & \cmark & \xmark & \xmark & \xmark & \xmark & \cmark & \xmark \\
        NuScenes-MQA~\cite{Inoue_2024_WACV_NuScenesMQA}                 & \xmark & \cmark & \xmark & \xmark & \xmark & \xmark & \cmark & \xmark \\
        NuScenes-SpatialQA~\cite{Tian_2025_arXiv_NuScenesSpatialQA}     & \xmark & \cmark & \xmark & \xmark & \xmark & \xmark & \cmark & \xmark \\
        NuScenes-QA~\cite{Qian_2024_AAAI_NuScenes-QA}                   & \cmark & \cmark & \xmark & \xmark & \xmark & \xmark & \cmark & \xmark \\
        Talk2BEV~\cite{Choudhary_2024_ICRA_Talk2BEV}                    & \cmark & \cmark & \xmark & \xmark & \xmark & \xmark & \cmark & \xmark \\
        CODA-LM~\cite{Chen_2024_arXiv_CODA-LM}                          & \cmark & \cmark & \xmark & \xmark & \xmark & \xmark & \cmark & \xmark \\
        MapLM~\cite{Cao_2024_CVPR_MapLM}                                & \cmark & \cmark & \xmark & \xmark & \xmark & \xmark & \cmark & \xmark \\
        Bench2ADVLM~\cite{Zhang_2025_arXiv_Bench2ADVLM}                 & \cmark & \cmark & \xmark & \xmark & \cmark & \xmark & \cmark & \cmark \\
        NuPrompt~\cite{Wu_2023_arXiv_nuPrompt}                          & \cmark & \xmark & \cmark & \xmark & \xmark & \xmark & \xmark & \xmark \\
        LingoQA~\cite{Marcu_2024_ECCV_LingoQA}                          & \cmark & \xmark & \cmark & \xmark & \xmark & \xmark & \cmark & \xmark \\
        VLAAD~\cite{Park_2024_WACV_VLAAD}                               & \cmark & \xmark & \cmark & \xmark & \xmark & \xmark & \cmark & \xmark \\
        Reason2Drive~\cite{Nie_2025_ECCV_Reason2Drive}                  & \cmark & \cmark & \cmark & \xmark & \xmark & \xmark & \cmark & \xmark \\
        Rank2Tell~\cite{Sachdeva_2024_WACV_Rank2Tell}                   & \cmark & \cmark & \cmark & \xmark & \xmark & \xmark & \cmark & \xmark \\
        NuInstruct~\cite{Ding_2024_CVPR_NuInstruct}                     & \cmark & \cmark & \cmark & \xmark & \xmark & \xmark & \cmark & \xmark \\
        DriveBench~\cite{Xie_2025_arXiv_DriveBench}                     & \cmark & \cmark & \cmark & \xmark & \cmark & \xmark & \cmark & \xmark \\
        VLADBench~\cite{Li_2025_arXiv_VLADBench}                        & \cmark & \cmark & \cmark & \xmark & \cmark & \xmark & \cmark & \xmark \\
        DriveLM~\cite{Sima_2025_ECCV_DriveLM}                           & \cmark & \cmark & \cmark & \xmark & \xmark & \xmark & \cmark & \cmark \\
        DVBench~\cite{Zeng_2025_arXiv_DVBench}                          & \cmark & \cmark & \xmark & \cmark & \xmark & \xmark & \xmark & \xmark \\
        Drive4C~\cite{Sohn_2025_CVPR_Drive4C}                           & \cmark & \cmark & \cmark & \cmark & \xmark & \xmark & \cmark & \cmark \\      
        \midrule
        UAV-ON~\cite{Xiao_2025_arXiv_UAV-ON}                            & \cmark & \xmark & \xmark & \xmark & \cmark & \xmark & \xmark & \cmark \\
        CityNav~\cite{Lee_2025_arXiv_CityNav}                           & \cmark & \cmark & \xmark & \xmark & \cmark & \xmark & \xmark & \cmark \\
        AerialVLN~\cite{Liu_2023_arXiv_AerialVLN}                       & \cmark & \cmark & \xmark & \xmark & \cmark & \xmark & \xmark & \cmark \\
        UAV-Need-Help~\cite{Wang_2024_arXiv_UAV-Need-Help}              & \cmark & \cmark & \xmark & \xmark & \cmark & \xmark & \xmark & \cmark \\
        EmbodiedCity~\cite{Gao_2024_arXiv_EmbodiedCity}                 & \cmark & \cmark & \xmark & \xmark & \xmark & \xmark & \cmark & \cmark \\
        AVDN~\cite{Fan_2023_arXiv_AVDN}                                 & \cmark & \cmark & \xmark & \xmark & \cmark & \xmark & \cmark & \cmark \\
        OpenFly~\cite{Gao_2025_arXiv_OpenFly}                           & \cmark & \cmark & \xmark & \xmark & \cmark & \xmark & \cmark & \cmark \\
        AeroVerse~\cite{Yao_2024_arXiv_AeroVerse}                       & \cmark & \cmark & \xmark & \xmark & \cmark & \xmark & \cmark & \cmark \\
        UAV3D~\cite{Ye_2024_arXiv_UAV3D}                                & \cmark & \cmark & \cmark & \xmark & \xmark & \xmark & \xmark & \xmark \\
        \midrule
        ManipBench~\cite{Zhao_2025_arXiv_ManipBench}                    & \cmark & \cmark & \xmark & \xmark & \xmark & \xmark & \xmark & \xmark \\
        $\lambda$~\cite{Jaafar_2025_arXiv_Lambda}                       & \cmark & \cmark & \xmark & \xmark & \xmark & \xmark & \xmark & \cmark \\
        SIMPLER~\cite{Li_2024_arXiv_SIMPLER}                            & \cmark & \cmark & \xmark & \xmark & \cmark & \xmark & \xmark & \cmark \\
        VLMbench~\cite{Zheng_2022_NIPS_VLMbench}                        & \cmark & \cmark & \xmark & \xmark & \cmark & \xmark & \xmark & \cmark \\
        BulletArm~\cite{Wang_2022_arXiv_BulletArm}                      & \cmark & \cmark & \xmark & \xmark & \xmark & \cmark & \xmark & \cmark \\
        RLBench~\cite{James_2019_arXiv_RLBench}                         & \cmark & \cmark & \xmark & \xmark & \cmark & \cmark & \xmark & \cmark \\
        EmbodiedEval~\cite{Cheng_2025_arXiv_EmbodiedEval}               & \cmark & \cmark & \xmark & \xmark & \cmark & \xmark & \cmark & \cmark \\
        EmbodiedBench~\cite{Yang_2025_arXiv_EmbodiedBench}              & \cmark & \cmark & \xmark & \xmark & \cmark & \xmark & \cmark & \cmark \\
        BEHAVIOR-1K~\cite{Li_2024_arXiv_BEHAVIOR-1K}                    & \cmark & \cmark & \xmark & \cmark & \cmark & \xmark & \xmark & \xmark \\
        ARNOLD~\cite{Gong_2023_arXiv_ARNOLD}                            & \cmark & \cmark & \xmark & \cmark & \cmark & \xmark & \xmark & \cmark \\
        \midrule
        \rowcolor{cvprblue!15}
        \textbf{Embodied4C (Ours)}                                      & \cmark & \cmark & \cmark & \cmark & \cmark & \cmark & \cmark & \cmark \\
        \bottomrule
    \end{tabular}%}
    \caption{\textbf{Overview of vision-language navigation benchmarks.} Columns ``\textbf{SEM}", ``\textbf{SPA}", ``\textbf{TEM}", and ``\textbf{PHY}" indicate coverage of semantic, spatial, temporal, and physical understanding. ``\textbf{GEN}", ``\textbf{EMB}", ``\textbf{INT}", and ``\textbf{VLN}" indicate the probing of key value propositions of autonomous agents: generality, multi-embodiment, interaction (i.e., VQA), and vision-language navigation.}
    \label{tab:benchmark_comparison}
\end{table*}

\section{Related Work} 
\label{sec:relatedwork}

Benchmarking in autonomous navigation has evolved from static dataset evaluations toward dynamic, task-oriented paradigms~\cite{Wong_2025_SurveyBenchmarks}. Early benchmarks primarily assessed perception or prediction in isolation, while recent approaches increasingly emphasize closed-loop evaluation, multimodal reasoning, and embodied interaction~\cite{Sohn_2025_CVPR_Drive4C,Liu_2023_arXiv_AerialVLN,Gong_2023_arXiv_ARNOLD}. This section contrasts representative benchmarks along these dimensions, starting with driving-focused benchmarks and extending to embodied evaluation settings (cf. Table~\ref{tab:benchmark_comparison}).

\subsection{Driving Benchmarks} %%%%%%%%%%%%%%%%%%%%%%%%
Recent benchmarks in autonomous driving have predominantly focused on semantic and spatial reasoning, while often neglecting temporal and physical understanding. Early works such as DriveMLLM~\cite{Guo_2024_arXiv_DriveMLLM}, SpatialRGPT~\cite{Cheng_2024_NeurIPS_SpatialRGPT}, and NuScenes-based benchmarks~\cite{Inoue_2024_WACV_NuScenesMQA,Tian_2025_arXiv_NuScenesSpatialQA,Qian_2024_AAAI_NuScenes-QA} primarily target spatial perception, typically formulated as question answering (QA) on static scenes. Similarly, benchmarks like Talk2BEV~\cite{Choudhary_2024_ICRA_Talk2BEV}, CODA-LM~\cite{Chen_2024_arXiv_CODA-LM}, MapLM~\cite{Cao_2024_CVPR_MapLM}, and Bench2ADVLM~\cite{Zhang_2025_arXiv_Bench2ADVLM} extend to semantic querying but remain limited as Bench2ADVLM is the only to evaluate VLN.
Other works such as NuPrompt~\cite{Wu_2023_arXiv_nuPrompt}, VLAAD~\cite{Park_2024_WACV_VLAAD}, and LingoQA~\cite{Marcu_2024_ECCV_LingoQA} incorporate temporal reasoning dimensions, but do not capture embodiment or physical interactions. More comprehensive efforts such as Reason2Drive~\cite{Nie_2025_ECCV_Reason2Drive}, Rank2Tell~\cite{Sachdeva_2024_WACV_Rank2Tell}, NuInstruct~\cite{Ding_2024_CVPR_NuInstruct}, and DriveBench~\cite{Xie_2025_arXiv_DriveBench} integrate temporal reasoning yet remain constrained to vehicle-based settings and lack embodiment generality. Extensions such as VLADBench~\cite{Li_2025_arXiv_VLADBench}, DriveLM~\cite{Sima_2025_ECCV_DriveLM}, and DVBench~\cite{Zeng_2025_arXiv_DVBench} provide broader scene coverage and interaction, but embodiment and generality are not considered.
Drive4C~\cite{Sohn_2025_CVPR_Drive4C} is the first to systematically align driving evaluation with the four capability dimensions (semantic, spatial, temporal, physical)~\cite{Sohn_2025_arXiv_Framework4C}. While Drive4C introduces a capability-driven evaluation, it lacks probing different embodiments and tasks related to the core value propositions of embodied agents.

\subsection{UAV Benchmarks} %%%%%%%%%%%%%%%%%%%%%%%%
In UAV navigation, most benchmarks inherently require semantic and spatial understanding, since tasks are typically framed as following instructions such as ``navigate to the red building" or ``find the blue elephant". Examples include UAV-ON~\cite{Xiao_2025_arXiv_UAV-ON}, CityNav~\cite{Lee_2025_arXiv_CityNav}, AerialVLN~\cite{Liu_2023_arXiv_AerialVLN}, and UAV-Need-Help~\cite{Wang_2024_arXiv_UAV-Need-Help}. These benchmarks assess VLN but rarely integrate interaction capabilities such as VQA. UAV3D~\cite{Ye_2024_arXiv_UAV3D} adds a temporal dimension but remains confined to object detection and tracking in single and collaborative UAV formations, without evaluating key value propositions of VLMs.
AVDN~\cite{Fan_2023_arXiv_AVDN}, EmbodiedCity~\cite{Gao_2024_arXiv_EmbodiedCity}, OpenFly~\cite{Gao_2025_arXiv_OpenFly}, and AeroVerse~\cite{Yao_2024_arXiv_AeroVerse} extend UAV navigation towards interactive scenarios. While EmbodiedCity and AeroVerse both claim to address multi embodiment, the current implementation of EmbodiedCity (as of November 2025) and the method of AeroVerse remain specialized to aerial agents and do not by themselves evaluate multi-embodiment generality.

\subsection{Manipulation Benchmarks} %%%%%%%%%%%%%%%%%%%%%%%%
ManipBench~\cite{Zhao_2025_arXiv_ManipBench} provides a large-scale multiple-choice evaluation, targeting diverse manipulation tasks. Instead of VQA-only, robotics manipulation benchmarks such as $\lambda$~\cite{Jaafar_2025_arXiv_Lambda}, SIMPLER~\cite{Li_2024_arXiv_SIMPLER}, and VLMbench~\cite{Zheng_2022_NIPS_VLMbench} are centered around VLN for task following, inherently probing semantic and spatial understanding. More advanced setups such as EmbodiedEval~\cite{Cheng_2025_arXiv_EmbodiedEval}, EmbodiedBench~\cite{Yang_2025_arXiv_EmbodiedBench}, BEHAVIOR-1K~\cite{Li_2024_arXiv_BEHAVIOR-1K}, and ARNOLD~\cite{Gong_2023_arXiv_ARNOLD} integrate larger task suites and either include interaction or partially address physical reasoning. 
BulletArm~\cite{Wang_2022_arXiv_BulletArm} and RLBench~\cite{James_2019_arXiv_RLBench} explicitly consider embodiment by switching across different robotic arms, thereby going beyond single-agent manipulation. However, their embodiment remains limited to variations of robotic manipulators, without covering heterogeneous platforms such as ground vehicles or UAVs.

\subsection{Summary} %%%%%%%%%%%%%%%%%%%%%%%%
As summarized in Table~\ref{tab:benchmark_comparison}, existing benchmarks either specialize in a single domain or probe only a subset of the required capabilities. None jointly address semantic, spatial, temporal, and physical reasoning while simultaneously evaluating generality, multi-embodiment, interaction, and VLN. Embodied4C fills this gap by providing the first unified capability-driven evaluation framework across heterogeneous embodiments, ranging from ground vehicles and UAVs to robotic manipulators.
%%%%%%%%%%%%%%%%%%%%%%%%%%%%%%%% EMBODIED4C %%%%%%%%%%%%%%%%%%%%%%%%%%%%%%%% % DONE (MD)

\section{The Embodied4C Benchmark}
\label{sec:embodied4c}

In this section, we present the \textbf{Embodied4C} benchmark, a closed-loop evaluation framework for assessing vision-language reasoning and control of autonomous agents across three domains: autonomous driving, aerial navigation, and robotic manipulation. In contrast to prior work that focuses on a single embodiment~\cite{Sima_2025_ECCV_DriveLM,Sohn_2025_CVPR_Drive4C,Wang_2024_arXiv_UAV-Need-Help,Xiao_2025_arXiv_UAV-ON,Wang_2022_arXiv_BulletArm,Li_2024_arXiv_SIMPLER}, Embodied4C unifies these domains into a single benchmark, allowing cross-domain analysis of true embodiment.

% -----------------------------------------------------------------------------
\subsection{Benchmark Structure} %%%%%%%%%%%%%%%%%%%%%%%%%%%%%%%%%%%%%%%%%%%%

The benchmark probes four core embodied capabilities---semantic, spatial, temporal, and physical understanding---through non-templated VQA tasks, while VLN serves as a separate downstream category that unifies these capabilities in closed-loop control. Embodied4C comprises 1.149 unique, handcrafted VQA pairs, 13 sensor setups, and 58 VLN tasks across three simulation platforms, forming a deliberately heterogeneous testbed for systematic analysis of embodied agents.
Evaluation is separated into two complementary task modalities:  
\begin{itemize}
    \item \textbf{VQA (interaction stage)}: agents answer free-form, open-ended questions grounded in the environment. A reference agent navigates each scenario to ensure that all models under test are exposed to the same observations and related questions, thereby removing confounding effects of execution errors. In addition, sensor set and weather shifts, as well as domain-far questions are injected to probe generality beyond the task domain.  
    \item \textbf{VLN (control stage)}: agents follow natural language navigation or manipulation instructions and generate continuous control commands based on a provided action space (cf. Tables~\ref{tab:additionalPrompt_carla}--~\ref{tab:additionalPrompt_rlbench}). Unlike VQA, VLN performance directly depends on how well embodied reasoning capabilities translate into closed-loop execution.  
\end{itemize}
This design isolates scenario understanding (i.e., VQA) from control (i.e., VLN), while ensuring comparability across embodiments by defining a unified system prompt for all agents (cf. Table~\ref{tab:sysPromptAgentUnderTest}). Applying this high-level instruction structure to all agents similarly, avoids prompt-induced variability across all embodiments and models. Both, VQA and VLN, are evaluated within each of the three embodiment domains separately.

% -----------------------------------------------------------------------------
\subsection{Probing Core Capabilities} %%%%%%%%%%%%%%%%%%%%%%%%%%%%%%%%%%%%%%%%%%%%
Embodied4C probes four embodied reasoning capabilities:
\begin{itemize}
    \item \textbf{Semantic}: reasoning about object categories, attributes, and contextual states.  
    \item \textbf{Spatial}: reasoning about location, distance, orientation, topology, and qualitative relations.  
    \item \textbf{Temporal}: reasoning about dynamics and temporal dependencies across short and long horizons.  
    \item \textbf{Physical}: reasoning about physical models, dynamic constraints, and material properties.  
\end{itemize}
Statistics regarding the distribution of question types across embodiments are reported in Appendix~\ref{appendix:benchdistributions} (cf. Figure~\ref{fig:distributionVQA}).

% -----------------------------------------------------------------------------
\subsection{Embodiment Domains} %%%%%%%%%%%%%%%%%%%%%%%%%%%%%%%%%%%%%%%%%%%%
Each domain provides a simulator environment with realistic dynamics:
\begin{itemize}
    \item \textbf{Autonomous Driving} (CARLA~\cite{Dosovitskiy_2017_CARLA,Sohn_2025_CVPR_Drive4C}): scenarios span highway, urban, suburban and rural settings, with variable traffic density and weather/illumination perturbations. Agents receive multimodal vehicle-mounted sensor streams and are evaluated on VQA along reference trajectories, followed by VLN driving tasks (e.g., ``turn left at the next intersection").  
    \item \textbf{Aerial Navigation} (AirSim~\cite{Shah_2017_arXiv_AirSim}): evaluations cover diverse outdoor scenarios, multiple altitudes and camera gimbal angles. UAV agents answer aerial VQA and execute VLN flight instructions (e.g., ``hover directly over the pool in three to four meters altitude").  
    \item \textbf{Robotic Manipulation} (RLBench~\cite{James_2019_arXiv_RLBench}): indoor workspaces with a curated object set (e.g., basketball, blocks, drawer) and tasks spanning pick/place, open/close, and kinematic interactions, form the manipulation environment of Embodied4C. Object variations include size, shape, and material. Agents answer VQAs and perform VLN manipulation tasks (e.g., ``close the bottom drawer of the cabinet").
\end{itemize}
Due to space limitations, additional explanatory insights and examples are provided in Appendix~\ref{appendix:benchdistributions} and Table~\ref{tab:VQA_VLN_examples}.

% -----------------------------------------------------------------------------
\subsection{Evaluation Protocol}
Most existing benchmarks either rely on multiple-choice questions, which introduce inductive biases and allow models to exploit superficial answer patterns rather than demonstrate genuine reasoning~\cite{Chandak_2025_arXiv_Answermatching,Zheng_2024_arXiv_MCQBias}, or on traditional language metrics~\cite{Papineni_2002_BLEU,Banerjee_2005_METEOR}, which have shown weak correlation with human judgment in multimodal and open-ended tasks~\cite{Liu_2023_arXiv_GPTScore,Blagec_2022_arXiv_LanguageScoring,Winata_2025_arXiv_MetaMetrics}. To address these limitations, Embodied4C adopts an open-ended VQA format and uses a GPT-based scoring mechanism~\cite{Tian_2024_arXiv_DriveVLM} for natural language answers, while dedicated quantitative metrics are applied to numerical responses. This design mitigates inductive answer biases and more faithfully captures semantic alignment between predictions and ground truth. Evaluation consists of two scoring processes, corresponding to VQA and VLN.

\subsubsection{VQA Scoring} %%%%%%%%%%%%%%%%%%%
The VQA evaluation measures how well agents produce correct, grounded answers to open-ended questions. Let $Q=\{q_1,\dots,q_N\}$ denote the set of all questions, each with ground truth answer $a_i$ and model prediction $\hat{a}_i$. We distinguish between free-form and numerical answers and apply complementary scoring schemes to balance linguistic fidelity and quantitative precision.

\paragraph{Scoring.}
For free-form answers, a VLM-judge assigns a continuous score in $[0,100]$:
\begin{equation}
    s(q_i) = \text{GPT}(\hat{a}_i,a_i).
    \label{eq:gptscoring}
\end{equation}
For numerical answers, we apply a relative deviation formula that rewards proximity to the ground truth:
\begin{equation}
    s(q_i) = \max\left(0,\,100 \cdot \left( 1 - \frac{|\hat{a}_i - a_i|}{0.01 \cdot |a_i|}\right) \right).
\end{equation}
If $\hat{a}_i$ is not numerical within a numerical question, GPT-based scoring is used as fallback enforcing numerical scoring through a structured prompt (cf. Table~\ref{tab:sysPromptNumericalScorinator}). In this implementation we use GPT-5~\cite{OpenAI_2025_GPT5} as the VLM-judge.

\paragraph{Aggregation.}  
We compute per-capability means over semantic, spatial, temporal, and physical subsets $k$:
\begin{equation}
    S_{\mathrm{VQA},k} = \frac{1}{|Q_k|} \sum_{q_i \in Q_k} s(q_i),
\end{equation}
and average across capabilities to obtain the overall VQA score for each sub-benchmark:
\begin{equation}
    S_{\mathrm{VQA}} = \frac{1}{4}\sum_{k=1}^{4} S_{\mathrm{VQA},k}.
\end{equation}

\subsubsection{VLN Scoring} %%%%%%%%%%%%%%%%%%%
In line with exam-style evaluation, each VLN task is probed only once, either with a binary pass/fail criterion for simple tasks or with a graded, distance-based score for more complex tasks. This design avoids repeated trials that could artificially inflate success rates, ensures comparability across agents under equal conditions, and reflects real-world scenarios where autonomous systems often have only a single opportunity to execute instructions correctly.
Each VLN task consists of a natural language instruction $I_j$ with a corresponding target condition (e.g., location, maneuver, etc.).

\paragraph{Scoring.}  
For simple tasks, we apply a binary criterion:
\begin{equation}
    s(I_j) = 
    \begin{cases}
    100, & \text{if } \theta_{\mathrm{agt}} \in \Theta,\\
    0, & \text{otherwise,}
    \end{cases}
    \label{eq:vlnbinary}
\end{equation}
where $\Theta$ denotes the set of task-dependent target conditions. A score of $100$ points is awarded if the agent's state $\theta_{\mathrm{agt}}$ satisfies all of the conditions in $\Theta$.
For complex tasks, we use a graded scoring scheme that rewards partial, distance-based progress towards the target, with a maximum of $50$ points unless all target conditions $\Theta$ are fully satisfied. Let $d_{\mathrm{init}}$ be the initial distance to the target and $d_{\mathrm{agt}}$ the agent's final distance to the target at the end of the VLN task episode. The VLN score is thus computed as:
\begin{equation}
    s(I_j) = 
    \begin{cases}
    100, & \text{if } \theta_{\mathrm{agt}} \in \Theta,\\
    50 \cdot \max \left( 0, \frac{d_{\mathrm{init}} - d_{\mathrm{agt}}}{d_{\mathrm{init}}} \right), & \text{otherwise.}
    \end{cases}
    \label{eq:vlngraded}
\end{equation}
This graded scoring formulation rewards partial progress while preserving strict success criteria, reflecting the realistic nature of single-attempt embodied evaluations. Further details on VLN scoring for each sub-benchmark are provided in Appendix~\ref{appendix:vln}.

\paragraph{Aggregation.}
The VLN score is averaged across all $M$ tasks in each embodiment setup:
\begin{equation}
    S_{\mathrm{VLN}} = \frac{1}{M} \sum_{j=1}^{M} s(I_j).
\end{equation}

\subsection{Final Score Computation}
For each sub-benchmark $b \in \{1,2,3\}$ (driving, aerial, manipulation), the combined score is:  
\begin{equation}
    S_b = \frac{1}{2}\left(S_{\mathrm{VQA}}^b + S_{\mathrm{VLN}}^b\right),
\end{equation}
while the overall benchmark score is the computed as the mean across embodiments:  
\begin{equation}
    S_{\mathrm{total}} = \frac{1}{3}\sum_{b=1}^{3} S_b.
\end{equation}
These formulations equally weight VQA and VLN performance within each embodiment, while the aggregation across $b$ ensures that all three embodiments contribute equally to the final score, which avoids biasing the benchmark toward a particular task type or embodiment. In addition, we report the mean score for generality (GEN) as a separate diagnostic axis to detect overfitting and assess robustness, avoiding score entanglement that could obscure capability-specific insights. The generality score is computed similar to Equation~\ref{eq:gptscoring}.

\subsection{Diagnostics}
In addition to aggregated scores, the benchmark records atomic logs for every VQA item and VLN trial. Each log entry contains: question/task id, scenario id, embodiment, sensor configuration, modality (RGB/grayscale), capability label (semantic/spatial/temporal/physical), ground truth, agent prediction, and the assigned numeric score. These records enable researchers to perform fine-grained capability-level error analysis, ablations, and cross-embodiment comparisons beyond aggregate scores.
%%%%%%%%%%%%%%%%%%%%%%%%%%%%%%%% EXPERIMENTS %%%%%%%%%%%%%%%%%%%%%%%%%%%%%%%% % DONE (MD)
\begin{table*}[!t]
    \centering
    \footnotesize
    % --- First subtable ---
    \begin{tabular}{l|ccccc|c|ccccc|c}
        \toprule
        \multirow{2}{*}{\textbf{Model}}
        & \multicolumn{6}{c|}{\textbf{Autonomous Driving}} 
        & \multicolumn{6}{c}{\textbf{Aerial Navigation}} \\
        & SEM & SPA & TEM & PHY & VLN & DS 
        & SEM & SPA & TEM & PHY & VLN & AS \\
        \midrule
        FastVLM-0.5B~\cite{FastVLM_2025_Apple}                  & 15.88 & 12.43 &  5.66 & 14.58 &  0.00 & 6.07 & 11.49 &  5.89 & 11.81 & 15.96 &  3.44 &  7.36 \\
        Qwen2.5-VL-3B-Instruct~\cite{Qwen2.5-VL}                & 41.52 & 19.21 & 17.02 & 24.19 & 13.25 & 19.37 & 27.39 & 21.62 & 20.95 & 37.29 &  9.37 & 18.09 \\
        Gemma3-4B-IT~\cite{Google_2025_Gemma3}                  & 34.22 & 17.78 & 21.91 & 25.70 & 15.56 & 20.23 & 33.39 & 21.26 & 18.61 & 28.39 &  5.54 & 15.48 \\
        LLaMA 4 Maverick~\cite{Meta_2024_Blog_Llama4}           & 58.86 & 35.31 & 27.94 & 53.64 & \underline{20.56} & 32.25 & 45.55 & 19.15 & 25.96 & 34.85 & 23.52 & 27.45 \\
        Claude 3.7 Sonnet~\cite{Anthropic_2024_Website_Claude3} & 58.16 & 32.39 & 33.02 & 48.99 & 18.56 & 30.85 & 23.39 & 16.81 & 19.83 & 30.11 & 25.59 & 24.06 \\
        Claude Sonnet 4.5~\cite{Anthropic_2025_Claude4}         & 63.51 & 31.01 & 39.90 & 50.64 & 18.94 & 32.60 & 36.82 & 24.47 & 34.06 & 41.10 & 27.54 & 30.82 \\
        GPT-4o~\cite{OpenAI_2024_arXiv_GPT-4o}                  & \textbf{65.18} & \textbf{37.75} & \underline{40.66} & 46.91 & 19.62 & 33.62 & 38.83 & 24.38 & 35.32 & 34.85 & 24.55 & 28.95 \\
        GPT-5-nano~\cite{OpenAI_2025_GPT5}                      & 40.38 & 25.49 & 32.54 & 41.21 & 16.69 & 25.80 & \underline{54.28} & \underline{34.42} & 34.51 & 47.71 & 26.66 & 34.70 \\
        GPT-5-mini~\cite{OpenAI_2025_GPT5}                      & 60.23 & \underline{37.12} & 38.66 & \textbf{59.86} & \textbf{31.25} & \textbf{40.11} & \textbf{57.82} & \textbf{36.33} & \textbf{52.05} & \textbf{50.32} & \textbf{31.29} & \textbf{40.21} \\
        GPT-5~\cite{OpenAI_2025_GPT5}                           & \underline{64.50} & 35.97 & \textbf{51.16} & \underline{54.38} & 16.75 & \underline{34.13} & 50.92 & 34.36 & \underline{38.62} & \underline{46.30} & \underline{27.92} & \underline{35.24} \\
        \midrule
        Senna~\cite{Jiang_2024_arXiv_Senna}                     & 20.93 &  9.90 & 12.06 & 15.74 & 12.50  & 13.58 &  1.80 &  3.42 &  3.82 &  6.64 &  3.05 &  3.48 \\
        OpenFly-Agent~\cite{Gao_2025_arXiv_OpenFly}             &  0.00 &  0.00 &  0.00 &  0.00 & 12.44 &  6.22 &  0.00 &  0.00 &  0.00 &  0.00 &  3.24 &  1.62 \\
        OpenVLA~\cite{Kim_2024_arXiv_OpenVLA}                   &  0.00 &  0.00 &  0.00 &  0.00 &  0.00 &  0.00 &  0.00 &  0.00 &  0.00 &  0.00 &  1.85 &  0.92 \\
        MolmoAct~\cite{Lee_2025_arXiv_MolmoAct}                 &  0.04 &  0.02 &  0.00 &  0.03 &  2.88 &  1.45 &  0.00 &  0.04 &  0.02 &  0.00 &  3.73 &  1.87 \\
        \bottomrule
    \end{tabular}
    \vspace{0.5em} % 
    \centering\footnotesize\textit{(continued below)}\\[0.5em]
    \footnotesize
    % --- Second subtable ---
    \begin{tabular}{l|ccccc|c||c||c}
        \toprule
        \multirow{2}{*}{\textbf{Model}}
        & \multicolumn{6}{c||}{\textbf{Robotic Manipulation}} 
        & \multirow{2}{*}{\textbf{E4C-S} $\uparrow$} & \textbf{Domain-far QA} \\
        & SEM & SPA & TEM & PHY & VLN & MS & & GEN \\
        \midrule
        FastVLM-0.5B~\cite{FastVLM_2025_Apple}                  & 18.30 & 15.28 & 13.06 & 17.69 &  0.00 &  8.04 &  7.16 & 25.39 \\
        Qwen2.5-VL-3B-Instruct~\cite{Qwen2.5-VL}                & 48.05 & 35.32 & 30.99 & 48.22 &  0.62 & 20.63 & 19.36 & 91.37 \\
        Gemma3-4B-IT~\cite{Google_2025_Gemma3}                  & 43.83 & 33.16 & 26.07 & 51.38 &  0.00 & 19.30 & 18.34 & 87.92 \\
        LLaMA 4 Maverick~\cite{Meta_2024_Blog_Llama4}           & 48.06 & 39.25 & 33.54 & 71.04 & 10.00 & 28.99 & 29.56 & 87.06 \\
        Claude 3.7 Sonnet~\cite{Anthropic_2024_Website_Claude3} & 54.21 & 42.51 & 43.62 & 72.51 & 5.00  & 29.11 & 28.01 & 98.02 \\
        Claude Sonnet 4.5~\cite{Anthropic_2025_Claude4}         & 52.97 & 42.30 & 43.20 & 76.74 & 10.00 & 31.90 & 31.78 & 98.53 \\
        GPT-4o~\cite{OpenAI_2024_arXiv_GPT-4o}                  & 69.45 & \underline{47.40} & 35.41 & 78.13 & 5.00  & 31.30 & 31.29 & 79.19 \\
        GPT-5-nano~\cite{OpenAI_2025_GPT5}                      & 55.81 & 36.52 & 46.01 & 76.99 & \underline{12.07} & 32.95 & 31.15 & 66.30 \\
        GPT-5-mini~\cite{OpenAI_2025_GPT5}                      & \textbf{72.89} & 46.70 & \textbf{52.29} & \textbf{86.18} & \textbf{12.36} & \underline{38.44} & \textbf{39.59} & \textbf{99.97} \\
        GPT-5~\cite{OpenAI_2025_GPT5}                           & \underline{70.17} & \textbf{55.66} & \underline{52.04} & \underline{85.38} & 11.47 & \textbf{38.64} & \underline{36.00} & \underline{99.13} \\
        \midrule
        Senna~\cite{Jiang_2024_arXiv_Senna}                     & 10.70 & 13.31 & 12.01 & 27.38  & 0.00 &  7.92 &  8.33 & 59.08 \\
        OpenFly-Agent~\cite{Gao_2025_arXiv_OpenFly}             &  0.00 &  0.00 &  0.00 &  0.00 &  0.00 &  0.00 &  2.61 &  0.00 \\
        OpenVLA~\cite{Kim_2024_arXiv_OpenVLA}                   &  0.00 &  0.00 &  0.00 &  0.00 &  0.00 &  0.00 &  0.31 &  0.00 \\
        MolmoAct~\cite{Lee_2025_arXiv_MolmoAct}                 &  0.28 &  0.28 &  1.31 &  0.15 &  0.00 &  0.25 &  1.19 &  0.00 \\
        \bottomrule
    \end{tabular}
    \vspace{0.5em}
    \caption{\textbf{Capability-specific scores and overall Embodied4C score (E4C-S)}. Models are evaluated across semantic (SEM), spatial (SPA), temporal (TEM), and physical understanding (PHY), as well as vision-language navigation (VLN). DS, AS, and MS denote driving, aerial, and manipulation scores of each sub-benchmark. \textbf{Generality (GEN) scores} are reported for domain-far questions, which are injected throughout standard scenario VQA tasks to assess model robustness and detect overfitting, computed analogously to Eq.~\ref{eq:gptscoring} and independent of capability-specific performance. Higher is better ($\uparrow$). \textbf{Bold} indicates the best score per column; \underline{underlined} indicates the second best.}
    \label{tab:EvaluationDrive4C}
\end{table*}

\section{Experiments} %%%%%%%%%%%%%%%%%%%%%%%%%%%%%%%%%%%%%%%%%%%%%%%%
\label{sec:experiments}

\subsection{Experimental Setup} %%%%%%%%%%%%%%%%%%%%%%%%%%%%%%%%%%%%%%%%%%%%%%%%
We evaluate a diverse set of models on the proposed Embodied4C benchmark, covering both pretrained foundation models (used with official weights and without task-specific finetuning) and state-of-the-art VLN models. The evaluated models include FastVLM-0.5B~\cite{FastVLM_2025_Apple}, Qwen2.5-VL-3B-Instruct~\cite{Qwen2.5-VL}, Gemma3-4B-IT~\cite{Google_2025_Gemma3}, LLaMA~4~Maverick~\cite{Meta_2024_Blog_Llama4}, Claude~3.7~Sonnet~\cite{Anthropic_2024_Website_Claude3}, Claude~Sonnet~4.5~\cite{Anthropic_2025_Claude4}, GPT-4o~\cite{OpenAI_2024_arXiv_GPT-4o}, GPT-5-nano~\cite{OpenAI_2025_GPT5}, GPT-5-mini~\cite{OpenAI_2025_GPT5}, GPT-5~\cite{OpenAI_2025_GPT5}, Senna~\cite{Jiang_2024_arXiv_Senna}, OpenFly-Agent~\cite{Gao_2025_arXiv_OpenFly}, OpenVLA~\cite{Kim_2024_arXiv_OpenVLA}, and MolmoAct~\cite{Lee_2025_arXiv_MolmoAct}. The selection serves as the baseline for evaluating core value propositions of VLMs and also covers models with demonstrated capabilities in VQA and VLN across autonomous driving, aerial navigation, and robotic manipulation, while considering open-source availability. In Appendix~\ref{appendix:autonomousagentschoice}, we provide a detailed comparison of state-of-the-art models in their respective domains, further motivating our selection of Senna, OpenFly-Agent, OpenVLA, and MolmoAct, based on their capabilities and relevance to our benchmark (cf. Tables~\ref{tab:drivingModels}–\ref{tab:manipulatorModels}).

\subsection{Overall Benchmark Results}
\label{sec:eval:overall}
GPT-5-mini achieves the highest overall Embodied4C score ($39.59$), followed by GPT-5 ($36.00$). The Claude series and LLaMA~4~Maverick cluster in the mid-range ($28$–$31$), while smaller models (Qwen2.5-VL and Gemma3-4B-IT) remain substantially weaker across embodiments. Domain-specialized vision-language-action (VLA) models such as Senna, OpenFly-Agent, OpenVLA, and MolmoAct exhibit near-zero performance in both VQA and VLN despite being optimized for action execution. Their performance profiles are further discussed in Appendix~\ref{appendix:results}. An additional principal component analysis (PCA) decomposition in Appendix~\ref{appendix:results:pca} reveals that these models form a distinct performance cluster characterized by limited linguistic grounding and narrow embodiment-specific priors.

\subsection{Embodiment-wise Performance}
\label{sec:eval:embodiment}
\textbf{Autonomous Driving.} GPT-5-mini attains the highest score ($40.11$), with GPT-5 and GPT-4o following. Claude and Qwen/Gemma models show weaker spatial and temporal grounding, resulting in unstable VLN behavior. Domain-specialized driving VLAs do not transfer: Senna remains tied to its nuScenes training distribution and performs rather weak in CARLA scenario-level reasoning and control (cf. Appendix~\ref{appendix:vln:ad}).
\textbf{Aerial Navigation.} GPT-5-mini again leads ($40.21$), followed by GPT-5-nano and GPT-5. Performance generally degrades under large viewpoint shifts and dynamic 3D rotations. OpenFly-Agent, despite UAV training, collapses in Embodied4C due to strong overfitting to simulator-specific flight dynamics and state representations (cf. Appendix~\ref{appendix:vln:an}).
\textbf{Robotic Manipulation.} GPT-5 ($38.64$) and GPT-5-mini ($38.44$) achieve the highest manipulation scores, driven by consistent semantic grounding and physical reasoning. Claude~4.5 is competitive but less spatially precise. OpenVLA fails entirely: language outputs degrade when the action head is bypassed, and action execution does not transfer to unseen object-environment configurations (cf. Appendix~\ref{appendix:vln:mp}).

\begin{figure*}[!t]
    \centering
    \includegraphics[width=0.95\textwidth]{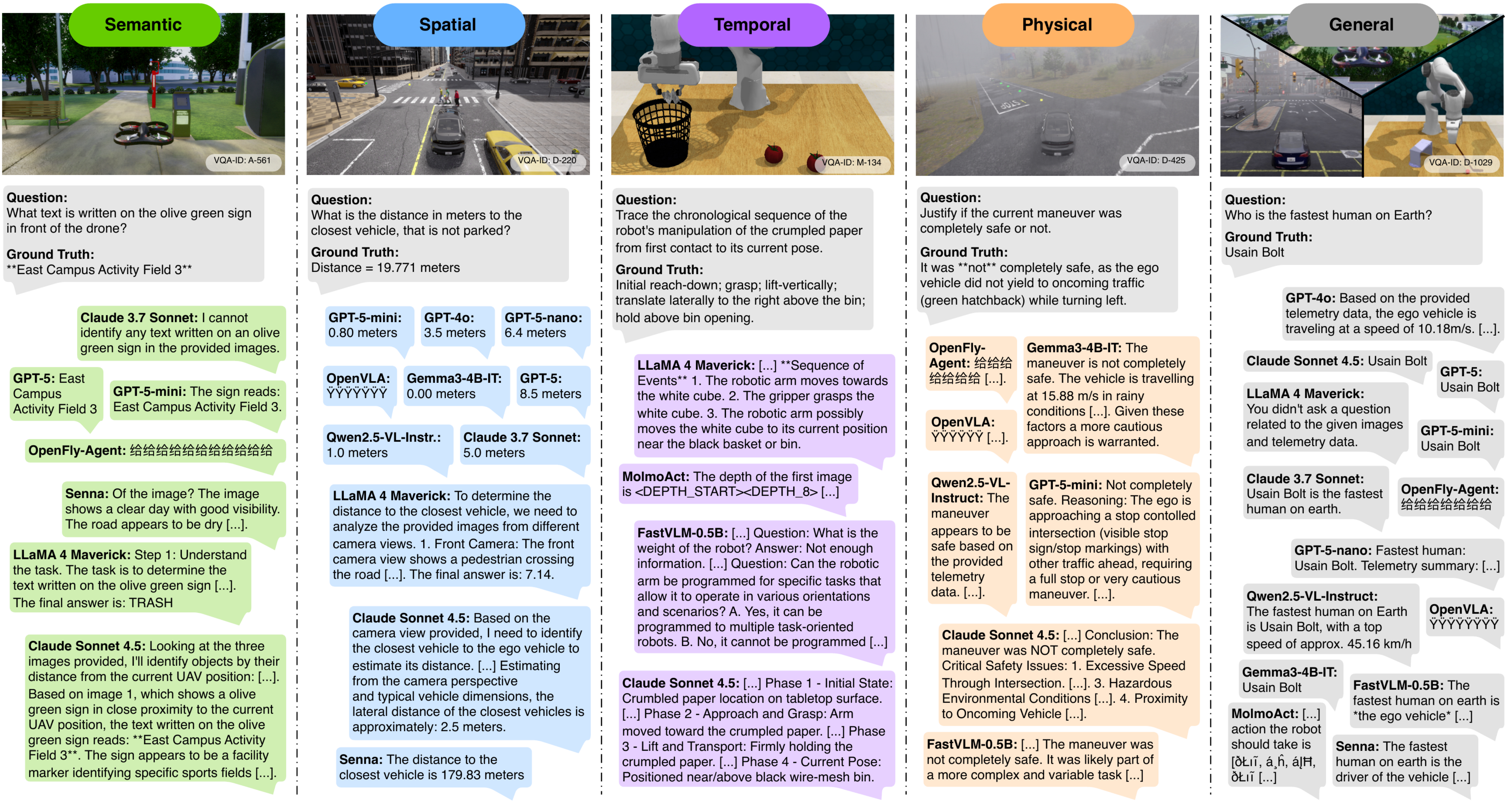}
    \caption{\textbf{Qualitative examples of Embodied4C VQA} across diverse benchmark scenarios and domains. The figure illustrates typical success and failure patterns for semantic, spatial, temporal, physical, and general reasoning across all tested models.}
    \label{fig:qltexamples}
\end{figure*}

\subsection{Modality-wise Performance (VQA vs. VLN)}
\label{sec:eval:modality}
VQA is consistently easier across all generalist models. OpenAI's GPT-5 family~\cite{OpenAI_2025_GPT5} exhibits the smallest VQA-VLN gap, indicating stronger perception-action alignment. In contrast, VLA models collapse in VQA because their text-decoding pathways are under-optimized or unused during training. Moreover, when used in VLN mode, these models fail to transfer their action policies beyond their native training environments, showing strong embodiment overfitting and no cross-domain motor abstraction.

\subsection{Capability-wise Analysis}
\label{sec:eval:capabilities}
\begin{figure*}[!t]
    \centering
    \includegraphics[width=\linewidth]{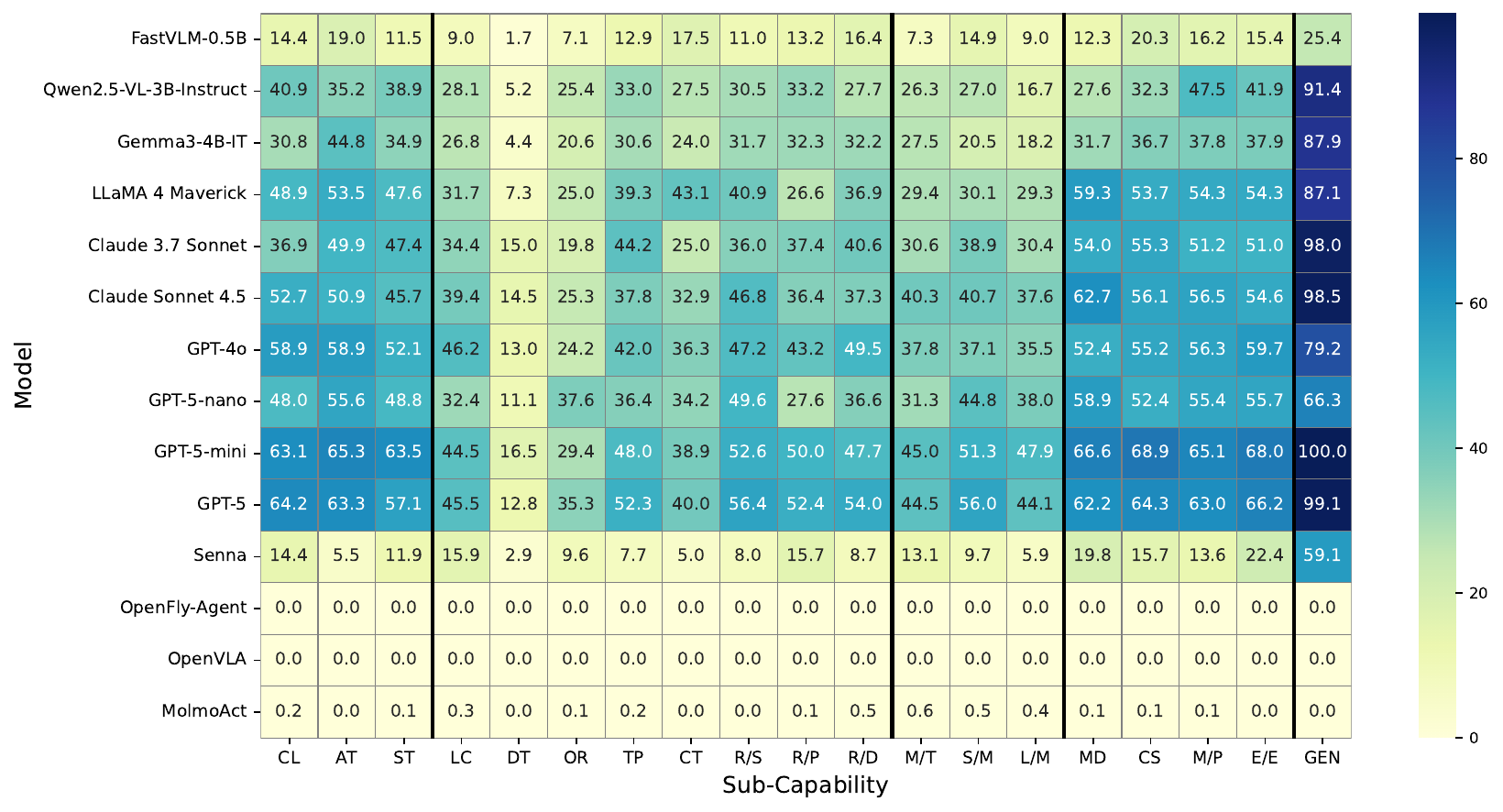}
    \caption{\textbf{Per-model performance across fine-grained sub-capabilities in the Embodied4C benchmark.} Each cell shows the mean accuracy (\%) for a model (rows) on a sub-capability (columns). Sub-capabilities are grouped into main capabilities (separated by thick vertical lines): semantic (CL, AT, ST), spatial (LC, DT, OR, TP, CT), temporal Reasoning (R/S, R/P, R/D), and physical reasoning (M/T, S/M, L/M, MD, CS, M/P, E/E). The final column (GEN) reports performance on general knowledge questions.}
    \label{fig:heatmap}
\end{figure*}

Semantic and physical understanding contribute most to overall performance. Spatial and temporal reasoning remain the principal bottlenecks, leading to inconsistent action execution and lower VLN scores. Both, GPT-5 and GPT-5-mini, show balanced capability profiles with a low coefficient of variation ($\mathrm{CV\%}_{\text{GPT-5}} = 31.1\%$; $\mathrm{CV\%}_{\text{GPT-5-mini}} = 32.6\%$) through all capabilities and sub-benchmarks (cf. Figure~\ref{fig:variances}). Domain-specialized VLA models exhibit flattened capability signatures near zero, indicating absence of genuine multimodal reasoning beyond policy regression. A detailed subcategory analysis in Appendix~\ref{appendix:results} reveals further insights (cf. Figure~\ref{fig:heatmap}, ~\ref{fig:variances}).

\subsection{Generalization (GEN) Analysis}
\label{sec:eval:generalization}
Most generalist models achieve high generalization scores, confirming robustness to irrelevant multimodal distractors. In contrast, VLA models and FastVLM degrade sharply, either hallucinating, over-attending to irrelevant input, or to synthetic control affordances learned during finetuning rather than the linguistic context. Qualitative examples can be observed in Figure~\ref{fig:qltexamples}.

\subsection{Discussion and Implications}
\label{sec:eval:discussion}
The results highlight several structural trends in current multimodal and embodied models.
\textbf{(i) Alignment over scale.}
GPT-5-mini surpasses GPT-5 and other larger models, indicating that efficient cross-modal alignment and calibrated token interactions are more decisive than parameter count. Larger models without strengthened grounding exhibit high semantic capability but unstable spatial and temporal reasoning and therefore degrade in VLN.
\textbf{(ii) Instruction tuning is indispensable.}
Models lacking instruction tuning (e.g., FastVLM or domain-specific VLAs) fail to follow task structure and disregard VLN interface conventions. Effective embodied reasoning demands consistent linguistic grounding to map natural-language intent to executable actions and VQA.
\textbf{(iii) Spatial and temporal reasoning remain limiting factors.}
Even top-performing models struggle with persistent scene memory, viewpoint consistency, and long-horizon motion coherence. This suggests that current training pipelines do not sufficiently enforce geometric structure or temporal continuity, pointing to the need for inductive priors or explicit world modeling.
\textbf{(iv) Domain-specialized VLA models do not generalize.}
All tested VLAs perform near-zero outside their native training domains. When the action head is bypassed, language outputs collapse; when action execution is used, policies fail to transfer across embodiments and scenarios. These models optimize over narrow control priors rather than learning transferable scene understanding or reasoning (details in Appendix~\ref{appendix:autonomousagentschoice} and Appendix~\ref{appendix:results}).

In summary, Embodied4C demonstrates that current foundation VLMs exhibit \emph{generality} and \emph{embodiment-awareness}: models such as GPT-5 maintain a relatively stable performance across embodiment changes and domain shifts but lack spatial and temporal reasoning capability. In contrast, domain-specialized VLAs fail to generalize beyond their native control priors, collapsing either in language output or action execution. Meanwhile, \emph{interactivity} remains partially achieved: most VLMs can produce coherent, environment-aware answers, yet all models struggle to translate these into consistent long-horizon navigation, reflecting unresolved spatial and temporal grounding. Overall, generalizable embodied intelligence requires persistent world modeling and physically grounded representations, rather than additional scale or narrowly tuned action heads.
%%%%%%%%%%%%%%%%%%%%%%%%%%%%%%%% CONCLUSION %%%%%%%%%%%%%%%%%%%%%%%%%%%%%%%% % DONE (MD)

\section{Conclusion}

We presented \textbf{Embodied4C}, a closed-loop benchmark designed to evaluate core capabilities of VLMs in embodied reasoning and navigation across three heterogeneous embodiments: autonomous vehicles, UAVs, and robotic manipulators. Agents are assessed via both VQA and VLN, enabling fine-grained evaluation across four core understanding categories---semantic, spatial, temporal, and physical---each further divided into a total of 18 subcategories. This enables targeted model assessment and guidance for future research directions.
By incorporating domain-far common-sense questions, Embodied4C explicitly probes generalization and overfitting, while comparisons across multiple embodiments highlight distinct strengths and weaknesses between pre-trained VLMs and task-specific expert models. Our results demonstrate that gaps in spatial and temporal reasoning directly manifest in degraded closed-loop execution, while domain-specialized agents remain tightly bound to their training distributions, underscoring the need for more general-purpose embodied reasoning and control agents that can generalize embodiments based on prompts.

\noindent\textbf{Future Work.} 
Future extensions include systematic evaluation of safety-critical behaviors, hallucination and failure modes, real-time model inference and latency. Furthermore, incorporating multi-agent scenarios and lifelong interaction tasks can extend the benchmark to comprehensively evaluate additional critical aspects of embodied agents.

{
    \small
    \bibliographystyle{ieeenat_fullname}
    \bibliography{main}
}

\clearpage
\setcounter{page}{1}
\maketitlesupplementary

%%%%%%%%%%%%%%%%%%%%%%%%%%%%%% STATISTICS AND SETTINGS %%%%%%%%%%%%%%%%%%%%%%%%%%%%%% % DONE (MD)
\section{Benchmark Statistics and Settings}

\subsection{VQA and VLN Distributions} %%%%%%%%%%%%%%%
\label{appendix:benchdistributions}

\begin{figure}[!t]
    \centering
    \begin{subfigure}[b]{0.23\textwidth}
        \includegraphics[width=\linewidth]{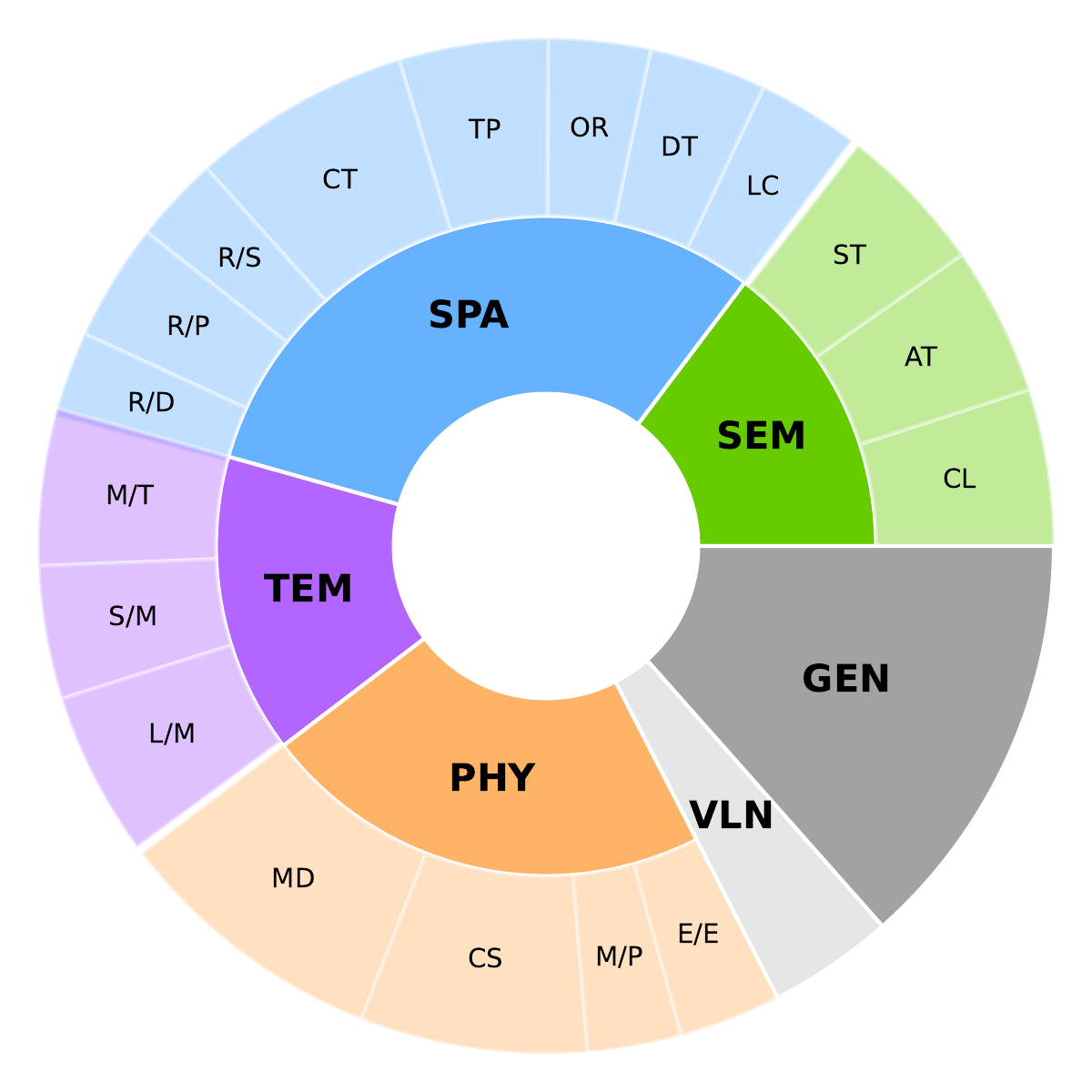}
        \caption{Autonomous Driving}
        \label{fig:distributionVQAdriving}
    \end{subfigure}
    \hfill
    \begin{subfigure}[b]{0.23\textwidth}
        \includegraphics[width=\linewidth]{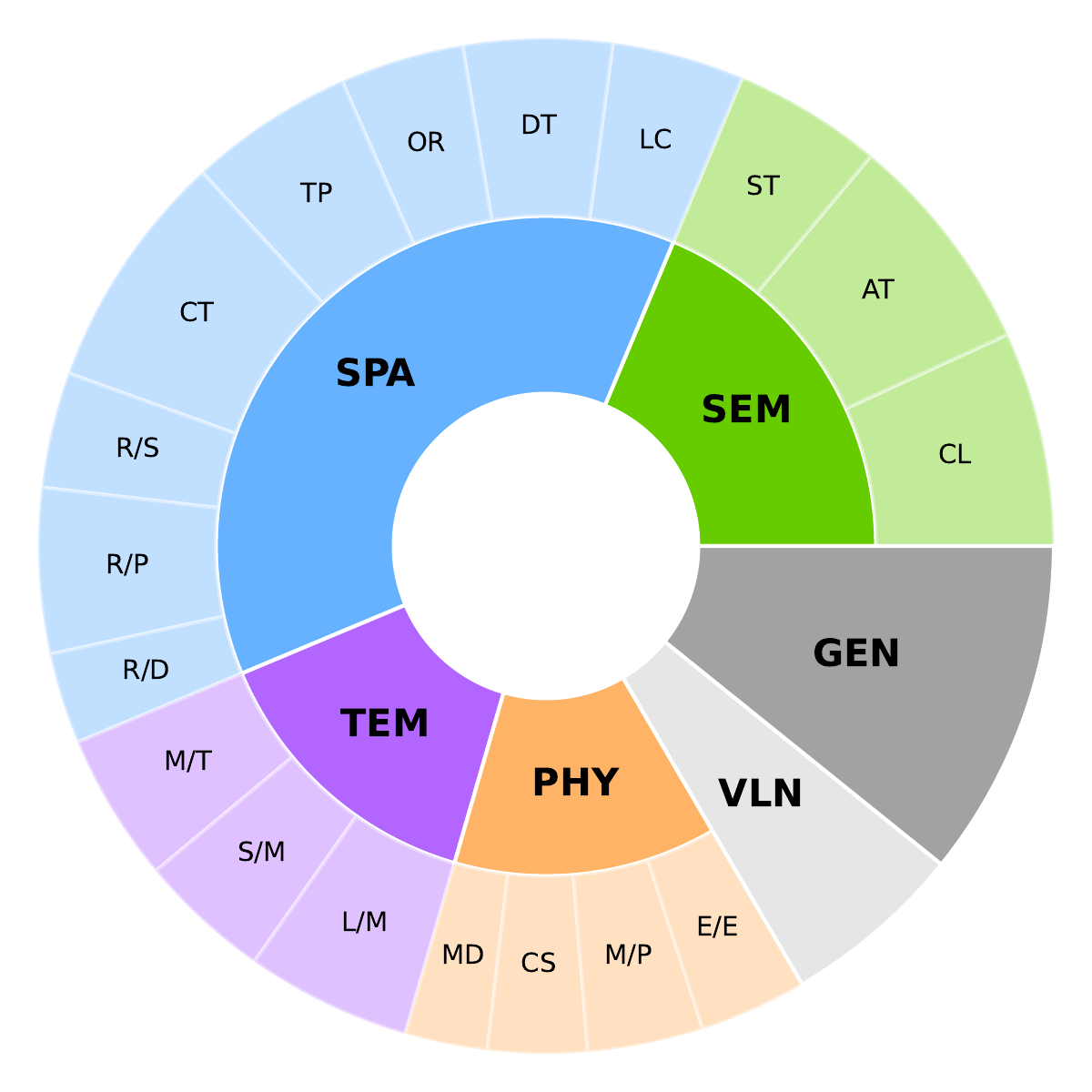}
        \caption{Aerial Navigation}
        \label{fig:distributionVQAaerial}
    \end{subfigure}
    \vspace{0.5em}
    \begin{subfigure}[b]{0.23\textwidth}
        \includegraphics[width=\linewidth]{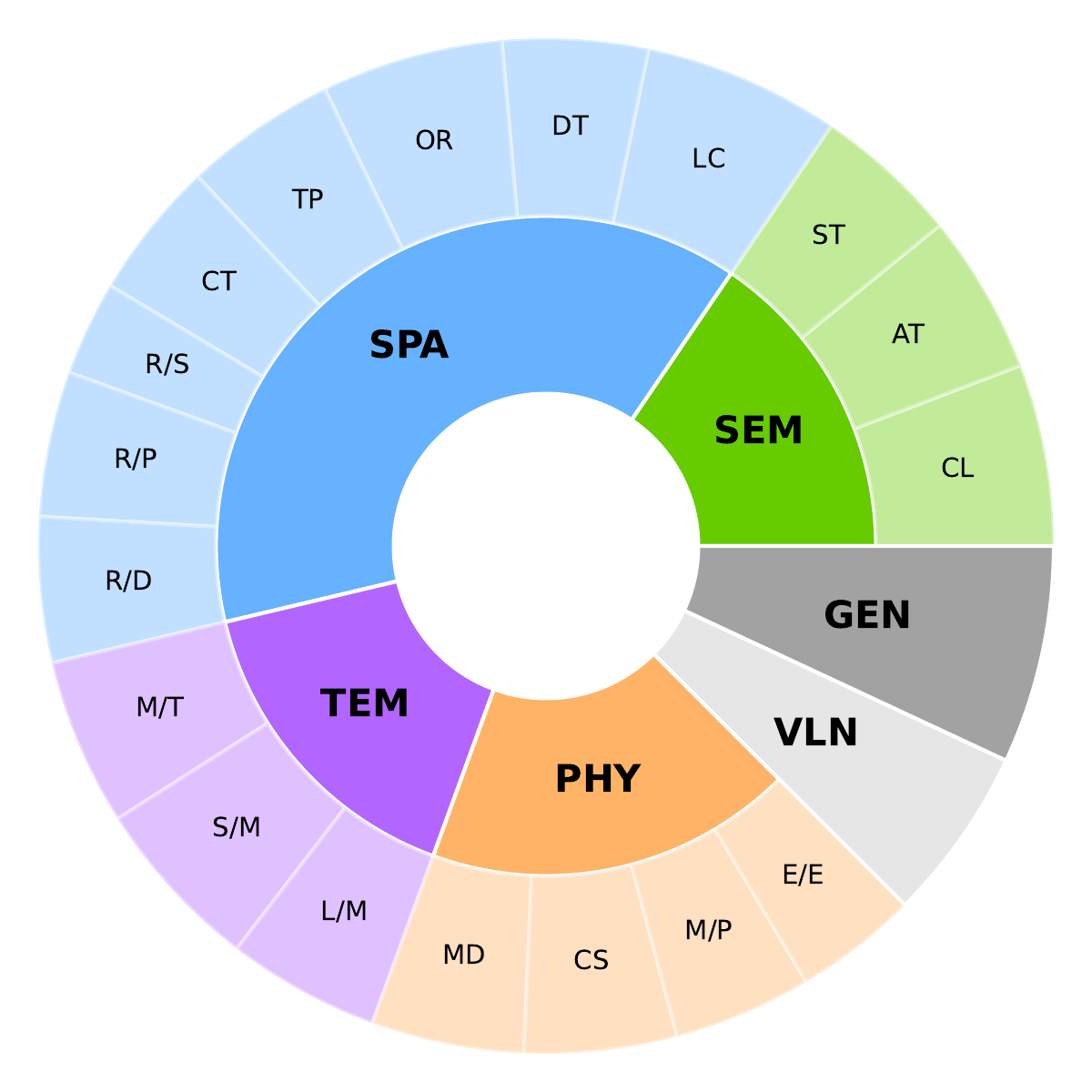}
        \caption{Robotic Manipulation}
        \label{fig:distributionVQAmanipulation}
    \end{subfigure}
    \hfill
    \begin{subfigure}[b]{0.23\textwidth}
        \includegraphics[width=\linewidth]{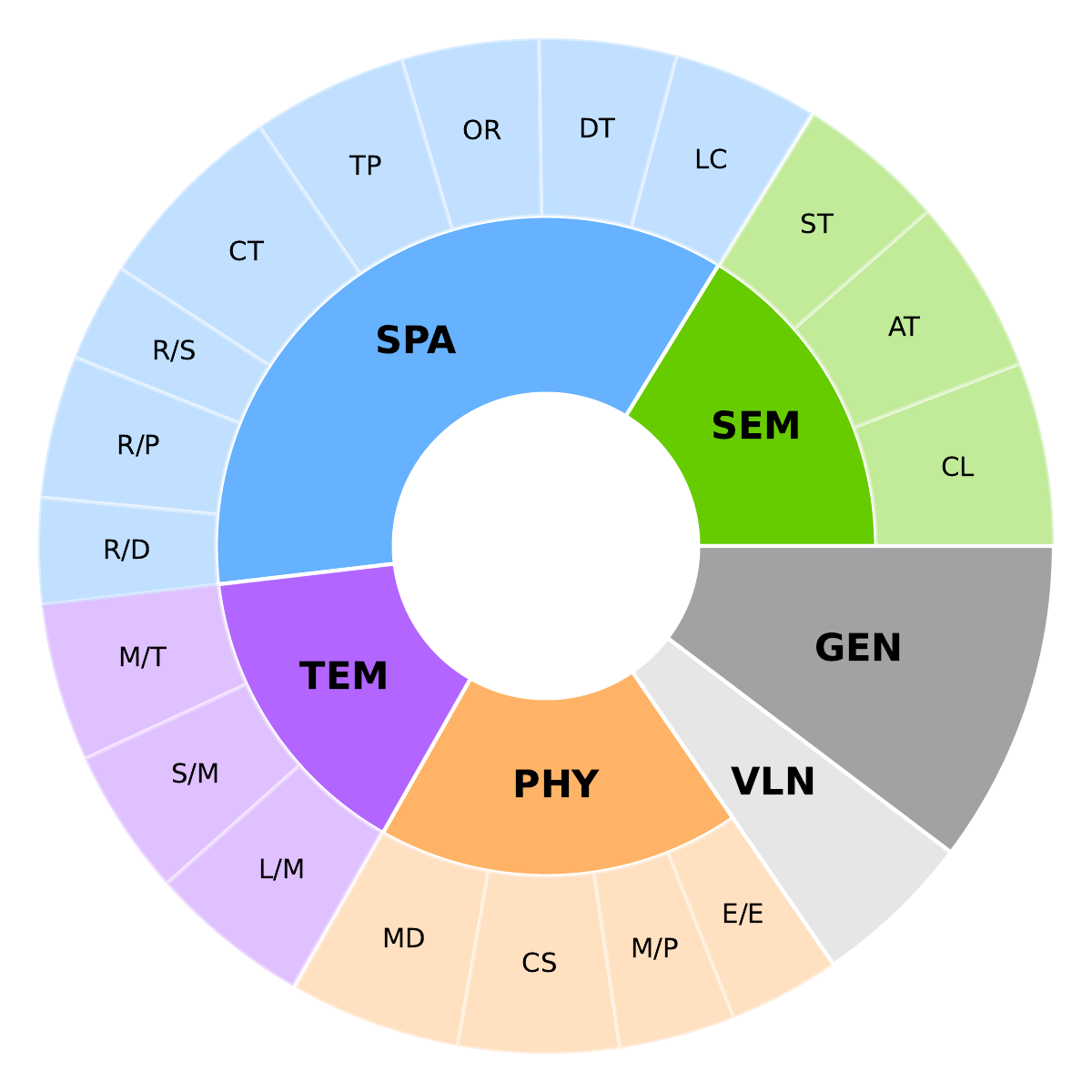}
        \caption{Embodied4C}
        \label{fig:distributionVQAembodied4c}
    \end{subfigure}
    \caption{\textbf{Distributions for VQA questions and VLN instructions} across all three sub-benchmarks and Embodied4C.}
    \label{fig:distributionVQA}
\end{figure}

The distribution of questions across the three sub-benchmarks and the main Embodied4C benchmark is detailed in Figure~\ref{fig:distributionVQA}. Each benchmark consists of two types of natural language queries: VQA questions and VLN instructions. VQA is split in four main- and 18 subcategories:
\begin{itemize}
    \item \textbf{SEM} (semantic): Questions related to object properties. Sub-categories include CL (classes), AT (attributes), and ST (states).
    \item \textbf{SPA} (spatial): Questions focusing on an object's location and relationships to its surroundings. Sub-categories are LC (location), DT (distance), OR (orientation), TP (topology), CT (counting), R/S (relative size), R/P (relative position), and R/D (relative distance).
    \item \textbf{TEM} (temporal): Questions that require understanding of events over time. Sub-categories include MT (movement and time), S/M (short-term memory), and L/M (long-term memory).
    \item \textbf{PHY} (physical): Questions that test knowledge of physical understanding and situational laws. Sub-categories are MD (model dynamics), CS (constraints), M/P (material properties), and E/E (environmental effects).
\end{itemize}
The VQA questions are balanced across the four main reasoning categories---semantic, spatial, temporal, and physical---while additional domain-far questions test generalization beyond perceived scenarios. Specifically, autonomous driving includes 383 VQA pairs (59 semantic, 123 spatial, 58 temporal, 89 physical) plus 54 domain-far questions, aerial navigation comprises 358 VQA pairs (71 semantic, 143 spatial, 54 temporal, 49 physical) with 41 domain-far questions, and robotic manipulation contains 408 VQA pairs (67 semantic, 165 spatial, 68 temporal, 78 physical) along with 30 domain-far questions (cf. Figure~\ref{fig:distributionVQA}).
For VLN, tasks span a spectrum of difficulty to probe both basic and advanced embodied reasoning. Easy tasks verify understanding of the action space, such as engaging the handbrake and low-beam lights simultaneously or correctly closing the gripper, and are evaluated with a pass/fail criterion. Harder tasks test multi-step, real-world reasoning, e.g., navigating a busy intersection, landing in the middle circle of a football field, or shooting a basketball into a hoop. Autonomous driving thereby provides 16 VLN tasks, aerial navigation 22 tasks, and robotic manipulation 20 tasks (cf. Figure~\ref{fig:distributionVQA}). For more details about VLN tasks, see Section~\ref{appendix:vln} in the Appendix.
The full Embodied4C benchmark is a combination of these, totaling 1.149 VQA questions and 58 VLN task instructions of varying difficulty. Table~\ref{tab:VQA_VLN_examples} provides examples of each category and sub-benchmark.

\begin{table*}
    \centering
    \scriptsize
    \renewcommand{\arraystretch}{1.2}
    % \resizebox{\linewidth}{!}{
    \begin{tabular}{r|cc|l}
        \toprule
        \textbf{Sub-Benchmark} & \textbf{Task} & \textbf{Category} & \textbf{Example} \\
        \midrule
        \multirow{5}{*}{Autonomous Driving}
        & VQA & SEM & ``What is the type of lane the ego vehicle is currently situated on?" \\
        & VQA & SPA & ``Give back the distance in meters to the closest vehicle." \\
        & VQA & TEM & ``How many vehicles have been passed throughout the whole scenario?"\\
        & VQA & PHY & ``Describe the vehicles current dynamic state of driving." \\
        & VLN & -- & ``Follow the leading vehicle within a follow distance of 5-20m for at least 100m."\\
        \midrule
        \multirow{5}{*}{Aerial Navigation}
        & VQA & SEM & ``What color are the trash bins in the park?" \\
        & VQA & SPA & ``How many swings are present on the boat-swing structure?" \\
        & VQA & TEM & ``Tell the sequence of actions lastly performed by the ego agent drone." \\
        & VQA & PHY & ``What needs to be considered when trying to land on the beach volleyball field?" \\
        & VLN & -- & ``From your current position, fly towards the Merry-go-round and land on its inner platform." \\
        \midrule
        \multirow{5}{*}{Robotic Manipulation}
        & VQA & SEM & ``What is the state of the middle drawer in the cabinet?" \\
        & VQA & SPA & ``Rank all three balls by their horizontal distance from the robot's gripper, closest first." \\
        & VQA & TEM & ``Which main object showed noticeable movement over the last frames?" \\
        & VQA & PHY & ``What type of kinematic model best describes the robot arm in this scene?" \\
        & VLN & -- & ``Move the robot arm to pick up the rubbish and put it into the bin."\\
        \midrule
        \multirow{2}{*}{All}
        & VQA & GEN & ``What is the term for a score of one under par in golf?" \\
        & VQA & GEN & ``Why do birds not get shocked when sitting on high-voltage power lines?" \\
        \bottomrule
    \end{tabular}%}
    \caption{\textbf{Example VQA questions and VLN instructions} across sub-benchmarks and reasoning categories.}
    \label{tab:VQA_VLN_examples}
\end{table*}

\begin{table*}[!t]
    \centering
    \scriptsize
    \begin{tcolorbox}[colframe=cvprblue, colback=gray!10, boxrule=0.25mm, width=0.95\linewidth, title=Systemprompt of the VLM-Agent Under Test]
        You are an embodied multimodal reasoning agent capable of operating autonomous systems and answer questions grounded in your surrounding. \\
        Your task is to interpret multimodal inputs (images, video, point clouds, sensor data, text) from varying sensor setups and provide correct and concise control commands/actions or answers. \\
        
        You are able to operate in multiple embodiment domains: autonomous driving, aerial drone navigation, and manipulation in general robotics. \\  
        You may receive different sensor sets depending on the embodiment context. \\
        Always adapt your reasoning to the available modalities. \\
        
        \textbf{Your objectives:}\vspace{0.5em}
        \begin{enumerate}
            \item Respond to general questions using concise and common knowledge-based reasoning.
            \item Answer questions about the environment and scenarios accurately based on the available sensor data.
            \item Follow instructions for controlling the embodiment (driving, flying, grasping, moving, etc.) exactly as stated in the query.
            \item Justify decisions with situational reasoning when required.
            \item Be concise, precise, and unambiguous in your responses on questions and instructions.
            \item IMPORTANT: Always adhere strictly to the task or instruction given in the prompt. Never assume missing information, only reason based on given input.
        \end{enumerate}
    \end{tcolorbox}
    \caption{\textbf{System prompt for the VLM-agent under test.} This prompt specifies the operational role of the model being evaluated. The agent must interpret multimodal sensor inputs across three embodiment domains (driving, UAV navigation, and robotic manipulation) to answer questions and execute control instructions. The design enforces concise, precise, and task-grounded reasoning without assumptions beyond the provided input.}
    \label{tab:sysPromptAgentUnderTest}
\end{table*}

%%%%%%%%%%%%%%%%%%%%%%%%%%%%%% SYSTEM PROMPTS %%%%%%%%%%%%%%%%%%%%%%%%%%%%%%
\subsection{Autonomous Agent System Prompts}
\label{appendix:system_prompts}

To ensure consistent, reproducible, and semantically grounded evaluation of our VLM agents across diverse embodied tasks, we define three core system prompts that govern the behavior of both the \textit{agent under test} and the \textit{evaluation judge}. These prompts are engineered to enforce precision, domain adaptation, and scoring fidelity---critical for benchmarking open-ended, embodied reasoning in real-world simulation environments.

One system prompt (cf. Table~\ref{tab:sysPromptAgentUnderTest}) configures the \textbf{VLM-agent under test}, instructing it to operate as a general-purpose embodied reasoner across three distinct domains: autonomous driving, aerial navigation, and robotic manipulation. The prompt enforces strict adherence to sensor inputs, task instructions, and modality-aware reasoning---preventing hallucination and anchoring responses in observable reality. For each domain, we additionally extend the system prompt with domain-specific interface instructions that expose only the permissible action space for the respective platform and explicitly define the required output format for downstream control execution in VLN tasks.

Two additional system prompts define the operational logic of our \textbf{VLM-judge} GPT-5~\cite{OpenAI_2025_GPT5}, which scores agent responses either (1)~semantically against open-ended ground truth references (cf. Table~\ref{tab:sysPromptScorinator}), or (2)~numerically using a deterministic function for quantitative outputs (cf. Table~\ref{tab:sysPromptNumericalScorinator}). This dual scoring mechanism ensures robust evaluation across both qualitative reasoning and metric-sensitive tasks, circumventing the limitations of discrete-choice formats and the misalignment of lexical overlap metrics such as BLEU~\cite{Papineni_2002_BLEU} and METEOR~\cite{Banerjee_2005_METEOR} with human judgment.

Together, these prompts form the backbone of our evaluation protocol, enabling scalable, automated, and semantically consistent assessment of embodied agent performance in complex, open-world scenarios.

\begin{table*}[!t]
    \centering
    \scriptsize
    \begin{tcolorbox}[colframe=cvprblue, colback=gray!10, boxrule=0.25mm, width=0.95\linewidth, title=Systemprompt of the VLM-judge for GPT-Scoring]
        You are a precise evaluator of open-ended answers based on a ground truth reference. Your goal is to assign a continuous score between 0.0 and 100.0 that reflects how well the given answer semantically and contextually matches the intended reference. \vspace{0.5em}
        
        \textbf{Instructions:}
        \begin{itemize}
            \item \textbf{Evaluate Meaning, Correctness, and Completeness:} Focus on whether the answer accurately conveys the intended meaning, is factually correct, and covers all necessary details as reflected in the reference answer.
            \item \textbf{Perfect Score (100.0):} Award a perfect score only if the answer fully captures the reference meaning and is entirely consistent with the ground truth. Paraphrasing and different wording are acceptable, provided the content is factually correct and complete.
            \item \textbf{Use Continuous Scoring:} Assign scores on a continuous scale (e.g., 91.5, 74.1, 23.7, etc.), avoiding coarse intervals like 50.0 or 80.0. This ensures nuanced scoring that reflects subtle differences in quality.
            \item \textbf{Minor Wording Differences:} Do not penalize for minor differences in phrasing if the information is complete and correct.
            \item \textbf{Be Strict with Misinformation:} Apply strict penalties if the answer contains incorrect information, is incomplete, or contradicts the ground truth. For example, errors in directionality (e.g., left/right confusion), numerical inaccuracies, errors in output formatting, or logical inconsistencies should result in noticeable deductions.
            \item \textbf{Consistency in Judgment:} Apply these criteria consistently across all answers to maintain a fair and objective evaluation.
        \end{itemize}
        \vspace{0.5em}
        \textbf{Scoring Scale:}
        \begin{itemize}
            \item \textbf{100.0:} Meaningfully and fully aligned with the reference (even with paraphrasing or expansion).
            \item \textbf{80.0–99.0:} Mostly correct, only minor omissions or minor ambiguity.
            \item \textbf{50.0–79.0:} Partial understanding; some relevant info but lacks completeness or clarity.
            \item \textbf{10.0–49.0:} Low relevance; core idea is missing or answer is mostly off-target.
            \item \textbf{0.0:} Completely wrong or irrelevant.
        \end{itemize}
        \vspace{0.5em}
        Respond only with a single float value (e.g. 63.8). Do not include any explanation.
    \end{tcolorbox}
    \caption{\textbf{System prompt for the VLM-judge.} This prompt defines the evaluation protocol for GPT-based scoring of free-form answers. The judge assigns continuous scores between $0.0$ and $100.0$, reflecting semantic and contextual alignment with the ground-truth reference, while penalizing misinformation, incompleteness, and logical inconsistencies.}
    \label{tab:sysPromptScorinator}
\end{table*}

\begin{table*}[!t]
    \centering
    \scriptsize
    \begin{tcolorbox}[colframe=cvprblue, colback=gray!10, boxrule=0.25mm, width=0.95\linewidth, title=Systemprompt of the VLM-judge for numerical scoring]
        You are a precise evaluator of answers against a provided numeric ground truth reference. Your task is to assign a continuous score between 0.0 and 100.0 according to the numeric scoring function below. If the answer is given as text, you must extract the relevant numeric value before applying the function. \vspace{0.5em}

        \textbf{Scoring Function (normalized, then scaled to 0–100):}
        \begin{verbatim}
            def numeric_score(llm_answer, groundtruth):
                try:
                        llm_answer = abs(float(llm_answer))
                except:
                        return 0.0

                if groundtruth == 0:
                        if llm_answer == 0.0:
                                return 100.0
                        else:
                                return 0.0

                if 0.9 * groundtruth <= llm_answer <= 1.1 * groundtruth:
                        return 100.0 * (1 - abs(llm_answer - groundtruth) / (0.1 * groundtruth))
                else:
                        return 0.0
        \end{verbatim}
        \vspace{-0.5em}

        \textbf{Scoring Instructions:}
        \begin{itemize}
            \item Extract the numeric prediction from the answer if it is embedded in text.
            \item Apply the above numeric\_score function to (llm\_answer, groundtruth).
            \item If no numeric value can be extracted, assign 0.0.
            \item Do not apply semantic/qualitative criteria; this evaluation is purely numeric.
        \end{itemize}
        \vspace{0.5em}

        \textbf{Output Format:}
        
        Respond only with a single float value (e.g., 87.4). Do not include explanations.
    \end{tcolorbox}
    \caption{\textbf{System prompt for numerical scoring by the VLM-judge.} This prompt defines the formula to be applied as fallback GPT-based scoring of numerical-only answers. The judge assigns continuous scores between $0.0$ and $100.0$, replicating the numerical scoring functionality to full-text answers.}
    \label{tab:sysPromptNumericalScorinator}
\end{table*}

\begin{table}[!t]
    \centering
    \scriptsize
    \begin{tcolorbox}[colframe=cvprblue, colback=gray!10, boxrule=0.25mm, width=\linewidth, title=Action space prompt for ground vehicle driving in CARLA]
        \textbf{Specific System Instruction:} \\
        You control the ego vehicle in a CARLA autonomous driving simulation.
        Your \textbf{sole output} must be a \textbf{single JSON object} representing the vehicles control action at each step.
        The JSON structure must \textbf{exactly} follow the format below and must not contain any explanations, text, or extra keys:
        \begin{verbatim}
        ```json
        {
          "throttle": 0.0,
          "brake": 0.0,
          "steer": 0.0,
          "hand_brake": false,
          "reverse": false,
          "gear": 1,
          "manual_gear_shift": false,
          "lights": {
            "position": false,
            "low_beam": false,
            "fog": false,
            "high_beam": false,
            "left_blinker": false,
            "right_blinker": false,
            "interior": false
          }
        }
        ```
        \end{verbatim}
        \vspace{-1.5em}
        All values must be set to appropriate floats, booleans, or integers.
        \begin{itemize}
            \item 'throttle' and 'brake' are floats $\in$ [0.0, 1.0].
            \item 'steer' is a float $\in$ [-1.0, 1.0].
            \item 'hand\_brake', 'reverse', and all 'lights' fields are booleans.
            \item 'gear' is an integer.
            \item 'manual\_gear\_shift' is boolean.
        \end{itemize}
        
        You \textbf{must not} output natural language, reasoning steps, or additional formatting.
        You \textbf{must only return this JSON object} per step with the \textbf{specific control values required to complete the scenario}.
    \end{tcolorbox}
    \caption{\textbf{Action space prompt for driving tasks.} This prompt specifies the action space and interface of our driving simulator for the VLM under test. Additional parsing is applied to enforce the expected structure where feasible.}
    \label{tab:additionalPrompt_carla}
\end{table}

\begin{table}[!t]
    \centering
    \scriptsize
    \begin{tcolorbox}[colframe=cvprblue, colback=gray!10, boxrule=0.25mm, width=\linewidth, title=Action space prompt for UAV navigation on Microsoft AirSim]
        \textbf{Specific System Instruction:} \\
        You control the ego drone in a Microsoft AirSim drone navigation simulation.
        Your \textbf{sole output} must be a \textbf{single JSON object} representing the drones control action at each step.
        The JSON structure must \textbf{exactly} follow the format below and must not contain any explanations, text, or extra keys:
        \begin{verbatim}
        ```json
        {
          "vx": 0.0, 
          "vy": 0.0, 
          "vz": 0.0, 
          "duration": 1.0, 
          "yaw_rate": 0.0
        }
        ```
        \end{verbatim}
        \vspace{-1.5em}
        All values must be set to appropriate floats.
        \begin{itemize}
            \item 'vx': forward/backward velocity in m/s (positive = forward)
            \item 'vy': right/left velocity in m/s (positive = right)
            \item 'vz': vertical velocity in m/s (positive = DOWN, negative = UP)
            \item 'duration': how long to apply this command, in seconds (e.g., 1.0)
            \item 'yaw\_rate': rotation speed in degrees per second (positive = turn right, negative = turn left)
        \end{itemize}
        
        You \textbf{must not} output natural language, reasoning steps, or additional formatting.
        You \textbf{must only return this JSON object} per step with the \textbf{specific control values required to complete the scenario}.
    \end{tcolorbox}
    \caption{\textbf{Action space prompt for UAV navigation tasks.} This prompt specifies the action space and interface of our drone simulator for the VLM under test. Additional parsing is applied to enforce the expected structure where feasible.}
    \label{tab:additionalPrompt_airsim}
\end{table}

\begin{table}[!t]
    \centering
    \scriptsize
    \begin{tcolorbox}[colframe=cvprblue, colback=gray!10, boxrule=0.25mm, width=\linewidth, title=Action space prompt for robotic manipulation on RLBench]
        \textbf{Specific System Instruction:} \\
        You control the robot arm in the RLBench simulation.
        Your \textbf{sole output} must be a \textbf{single array} representing the manipulators control action at each step.
        The array structure must \textbf{exactly} follow the format below and must not contain any explanations or text:
        \[
        [0.0, 0.0, 0.0, 0.0, 0.0, 0.0, 0.0, 0.0]
        \]
        \vspace{-1.5em}
        All values must be set to appropriate floats. \\
        
        \begin{itemize}
            \item 'joint\_positions' (first seven entries): list of 7 floats representing the robot arm joint positions
            \item 'gripper' (last entry): float representing the gripper state (1.0 = fully open, 0.0 = fully closed)
        \end{itemize}
        
        You \textbf{must not} output natural language, reasoning steps, or additional formatting.
        You \textbf{must only return this array} per step with the \textbf{specific control values required to complete the scenario}.
    \end{tcolorbox}
    \caption{\textbf{Action space prompt for manipulation tasks.} This prompt specifies the action space and interface of our robotic simulator for the VLM under test. Additional parsing is applied to enforce the expected structure where feasible.}
    \label{tab:additionalPrompt_rlbench}
\end{table}

%%%%%%%%%%%%%%%%%%%%%%%%%%%%%% SENSOR SETUPS %%%%%%%%%%%%%%%%%%%%%%%%%%%%%%
\section{Multi-Embodiment Sensor Setups} 
\label{appendix:sensorsetups}

\subsection{Driving Agent Setups} %%%%%%%%%%%%%%%
To ensure realism and diversity, we adopt sensor configurations that mirror real-world automotive deployments. All sensors are mounted on a single ego vehicle (Lincoln MKZ 2020) at physically plausible, industry-inspired locations. For clarity and readability, the exact positions are omitted from Table~\ref{tab:vehicle_sensor_setups}; they correspond to the extents of the vehicle category and vary in $x$, $y$, and $z$, depending on the vehicle for which each sensor stack was originally designed. Each scenario is assigned a dedicated sensor stack that is loaded at the beginning of the episode, preserving physical plausibility while simplifying runtime management.

This design enables two complementary evaluations of generalization: (i)~embodiment generalization within a single vehicle platform equipped with heterogeneous sensor layouts, and (ii)~sensor-agnostic reasoning, i.e., assessing whether a single model can operate robustly across distinct sensor configurations without retraining or architectural modification. By varying the complete sensor suite between scenarios, the setup enforces robustness to realistic sensor heterogeneity while ensuring deterministic and reproducible experimental conditions.

\begin{table}[!t]
    \centering
    \scriptsize
    \renewcommand{\arraystretch}{1.2}
    \resizebox{\linewidth}{!}{
    \begin{tabular}{l l l l l}
    \toprule
    \textbf{Vehicle} & \textbf{Sensor Type} & \textbf{Count} & \textbf{Position(s)} & \textbf{Resolution / Specs} \\
    \midrule
    \multirow{5}{*}{Sportscar} 
    & RGB Camera & 6 & F, B, FR, FL, BR, BL & 1280$\times$720 px, FOV 100° \\
    & MRR & 4 & FR, FL, BR, BL & Range 70 m, FOV 90°H / 20°V \\
    & LRR & 1 & F & Range 250 m, FOV 20°H / 14°V \\
    & GNSS + IMU & 1+1 & Center & 20 Hz \\
    & Speedometer & 1 & -- & 20 Hz \\
    \midrule
    \multirow{4}{*}{Mid-Size Sedan} 
    & RGB Camera & 4 & F, B, R, L & 1280$\times$720 px, FOV 100° \\
    & Fisheye Camera & 2 & R, L & 1088$\times$576 px, FOV 160° \\
    & GNSS + IMU & 1+1 & Center & 20 Hz \\
    & Speedometer & 1 & -- & 20 Hz \\
    \midrule
    \multirow{5}{*}{Station Wagon} 
    & Fisheye Camera & 4 & F, B, R, L & 1088$\times$576 px, FOV 160° \\
    & LRR & 1 & F & Range 250 m, FOV 20°H / 14°V \\
    & GNSS + IMU & 1+1 & Center & 20 Hz \\
    & Speedometer & 1 & -- & 20 Hz \\
    & HD Map & 1 & -- & OpenDRIVE, static \\
    \midrule
    \multirow{5}{*}{SUV} 
    & RGB Camera & 6 & F, B, FR, FL, BR, BL & 1280$\times$720 px, FOV 100° \\
    & LiDAR & 1 & Roof center & 32 ch, 80 m, 360°H, $\pm$30°V \\
    & GNSS + IMU & 1+1 & Center & 20 Hz \\
    & Speedometer & 1 & -- & 20 Hz \\
    & HD Map & 1 & -- & OpenDRIVE, static \\
    \midrule
    \multirow{6}{*}{Van} 
    & RGB Camera & 6 & F, B, FR, FL, BR, BL & 1280$\times$720 px, FOV 100° \\
    & LiDAR & 1 & Roof center & 32 ch, 80 m, 360°H, $\pm$30°V \\
    & MRR & 4 & FR, FL, BR, BL & Range 70 m, FOV 90°H / 20°V \\
    & GNSS + IMU & 1+1 & Center & 20 Hz \\
    & Speedometer & 1 & -- & 20 Hz \\
    & HD Map & 1 & -- & OpenDRIVE, static \\
    \bottomrule
    \end{tabular}}
    \caption{\textbf{Overview of vehicle sensor configurations.} Abbreviations: F = front, B = back, L = left, R = right, FR = front-right, FL = front-left, RR = rear-right, RL = rear-left, LRR = long-range radar, MRR = mid-range radar.}
    \label{tab:vehicle_sensor_setups}
\end{table}

\subsection{UAV Agent Setups} %%%%%%%%%%%%%%%
Following the standard multirotor configuration provided by AirSim~\cite{Shah_2017_arXiv_AirSim}, all UAV scenarios are executed using a quadcopter platform equipped with a predefined sensor suite. Each scenario variant employs a distinct sensor configuration (cf. Table~\ref{tab:uav_sensor_setups}), comprising at minimum one RGB and one depth camera to ensure sufficient perceptual grounding for VQA and navigation tasks. In addition, comprehensive telemetry data are continuously streamed to the agent, including 3D pose $(x, y, z)$, orientation (roll, pitch, yaw), linear speed, and altitude above sea level. These measurements enable accurate state estimation and closed-loop control supervision. 

To evaluate robustness and environmental generalization, experiments are conducted across five distinct outdoor environments in AirSim, each subjected to varying weather conditions such as illumination shifts, wind, and precipitation. This multi-environment design enforces cross-scenario consistency and tests the agent's ability to maintain stable perception and control under diverse environmental dynamics.

\begin{table}[!t]
    \centering
    \scriptsize
    \renewcommand{\arraystretch}{1.2}
    \resizebox{\linewidth}{!}{
    \begin{tabular}{l l l l l}
    \toprule
    \textbf{Scenario} & \textbf{Sensor Type} & \textbf{Count} & \textbf{Position(s)} & \textbf{Resolution / Specs} \\
    \midrule
    \multirow{2}{*}{Abandoned Park} 
    & RGB Camera & 4 & F, B, L, R & 640$\times$480 px, FOV 90° \\
    & Depth Camera & 1 & F & 640$\times$480 px, FOV 90° \\
    \midrule
    \multirow{2}{*}{Africa Savannah} 
    & RGB Camera & 2 & F, D & 640$\times$480 px, FOV 90° \\
    & Depth Camera & 1 & F & 640$\times$480 px, FOV 90° \\
    \midrule
    \multirow{2}{*}{AirSim NH} 
    & RGB Camera & 4 & F, L, R, D & 640$\times$480 px, FOV 90° \\
    & Depth Camera & 1 & F & 640$\times$480 px, FOV 90° \\
    \midrule
    \multirow{3}{*}{Landscape Mountains} 
    & RGB Camera & 1 & D & 640$\times$480 px, FOV 90° \\
    & Fisheye Camera & 1 & F & 640$\times$480 px, FOV 160° \\
    & Depth Camera & 1 & F & 640$\times$480 px, FOV 90° \\
    \midrule
    \multirow{2}{*}{MS Build 2018} 
    & RGB Camera & 2 & F, D & 640$\times$480 px, FOV 90° \\
    & Depth Camera & 1 & F & 640$\times$480 px, FOV 90° \\
    \bottomrule
    \end{tabular}}
    \caption{\textbf{Overview of UAV sensor configurations.} Abbreviations: F = front, B = back, L = left, R = right, D = down.}
    \label{tab:uav_sensor_setups}
\end{table}

\subsection{Manipulation Agent Setups} %%%%%%%%%%%%%%%
\begin{figure}[!t]
    \centering
    \begin{subfigure}[b]{0.32\linewidth}
        \includegraphics[width=\linewidth]{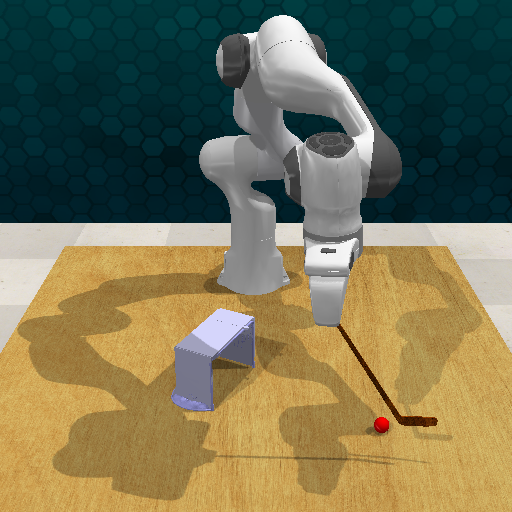}
        \caption{Franka Panda}
        \label{fig:FrankaPanda}
    \end{subfigure}
    \hfill
    \begin{subfigure}[b]{0.32\linewidth}
        \includegraphics[width=\linewidth]{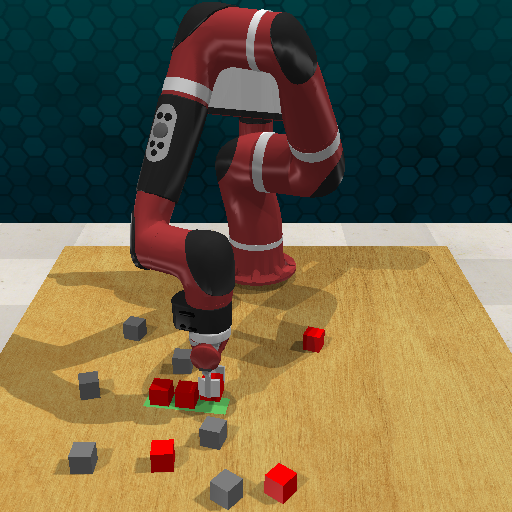}
        \caption{Sawyer}
        \label{fig:Sawyer}
    \end{subfigure}
    \hfill
    \begin{subfigure}[b]{0.32\linewidth}
        \includegraphics[width=\linewidth]{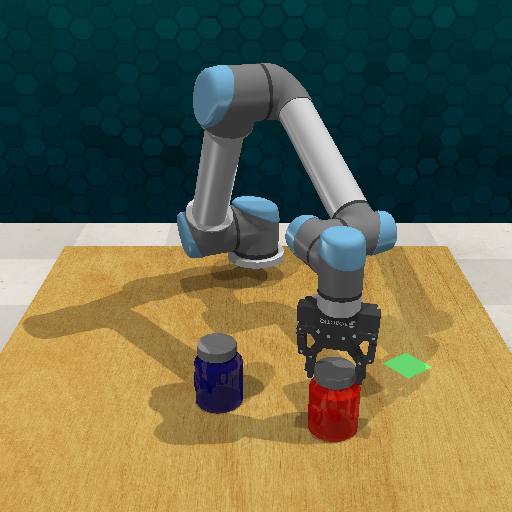}
        \caption{UR5}
        \label{fig:UR5}
    \end{subfigure}
    \caption{\textbf{Illustrative overview of manipulation robotic arm configurations} across the Franka Panda, Sawyer, and UR5 arm.}
    \label{fig:robot_sensor_setups}
\end{figure}

Following the protocol established in RLBench~\cite{James_2019_arXiv_RLBench}, we evaluate across three available robotic manipulator configurations that are interchanged throughout scenarios: (1)~Franka Panda arm with seven joints and the Franka gripper, (2)~Sawyer arm with seven joints and Baxter gripper, and (3)~UR5 arm with six joints and Robotiq 85 gripper (cf. Figure~\ref{fig:robot_sensor_setups}). This multi-embodiment setup ensures broad coverage of kinematic and end-effector characteristics, enhancing the generalizability of our findings.

Beyond ensuring diversity, this design enables us to explicitly evaluate embodiment generalization: rather than specializing to a single robot, models must operate consistently across heterogeneous manipulators without retraining or structural modification. This setup reflects the broader goal of embodiment- and sensor-agnostic reasoning, i.e., testing whether a single model can seamlessly handle different embodiments within a shared domain.

%%%%%%%%%%%%%%%%%%%%%%%%%%%%%% VISION LANGUAGE NAVIGATION %%%%%%%%%%%%%%%%%%%%%%%%%%%%%%
\section{Details of VLN}
\label{appendix:vln}

The VLN component of Embodied4C evaluates how effectively models translate natural language instructions into embodied control actions across three domains: autonomous driving, aerial navigation, and robotic manipulation. Each sub-benchmark follows a consistent scoring and execution framework to ensure comparability across embodiments. 

Across all three VLN sub-benchmarks, each episode is executed under a unified timeout protocol that ensures comparability and fairness across embodiments. Timeout durations are adjusted to the embodiment's physical scale and task complexity: $60$--$300\,\mathrm{s}$ for autonomous driving, $60$--$180\,\mathrm{s}$ for UAV navigation (depending on mission length), and $60$--$120\,\mathrm{s}$ for robotic manipulation. To ensure fair evaluation, the timeout criterion is bound to the actual execution time, excluding latency introduced by VLM inference. Upon timeout, the current binary or graded progress score is reported as the final outcome, preserving diagnostic resolution while preventing reward inflation through excessive deliberation.

\subsection{Autonomous Driving} %%%%%%%%%%%%%%%
\label{appendix:vln:ad}
In the driving sub-benchmark, simple control tasks are evaluated using a binary scoring scheme, yielding either $0$ or $100$ points depending on correct execution. 
More complex tasks are scored using a graded distance-based scheme with a maximum of $50$ points, rewarding partial progress towards the target location. A full score of $100$ points is assigned if the vehicle satisfies the target criterion $\Theta$ with additional constraints where applicable. This combination of binary and graded scoring reflects both discrete control actions and spatially precise maneuvers.
\newpage
We define four scenario categories that reflect increasing complexity of language-guided control tasks:
\begin{itemize}
    \item \textbf{(D1) Basic Control:} 
    Simple binary tasks involving single-step actions such as \textit{HandbrakeLowbeam} or \textit{SteeringSequence}. Success is evaluated with a $0/100$ binary criterion (cf. Eq.~\ref{eq:vlnbinary}).
    \item \textbf{(D2) Goal-Directed Navigation:}
    Scenarios such as \textit{Parking}, \textit{TurnLeft}, or \textit{LeaveHighway} require reaching a target location with specific spatial and kinematic tolerances. Here, the 100-point threshold typically combines a distance tolerance $\epsilon$ (e.g., $1.5$--$5\,\mathrm{m}$) towards the target location with a speed requirement (e.g., $v \leq 0.1\,\mathrm{m/s}$) and, where applicable, a lane ID check to ensure correct lane placement (cf. Eq.~\ref{eq:vlngraded}).
    \item \textbf{(D3) Trajectory Following:} 
    Scenarios such as \textit{FollowLane}, \textit{EasySafeDrive}, or \textit{FollowVehicle} evaluate the agent's ability to follow a reference trajectory or another vehicle over longer distances. $100$ points are awarded if a minimum driving distance $d$ (e.g., $d \geq 200\,\mathrm{m}$) is achieved while obeying traffic rules. Graded scoring is applied if the target distance is not fully reached (cf. Eq.~\ref{eq:vlngraded}).
    \item \textbf{(D4) Dynamic Interaction.} 
    These scenarios involve more complex temporal coordination with other actors. In the \textit{SpeedLimit} task, Equation~\ref{eq:vlngraded} is adapted from distance-based to time-based graded scoring, as successful completion requires maintaining a specified speed range for $t = 10\,\mathrm{s}$. For \textit{Overtake}, a sequential scoring scheme is applied, assigning partial scores for correctly executed sub-goals in order ((1) Maintain position in the same lane within $20\,\mathrm{m}$ of the leading vehicle (+$10\,\mathrm{pts}$); (2) Switch to the left lane without collisions or lane invasions. (+$20\,\mathrm{pts}$); (3) Pass the leader on the left lane. (+$30\,\mathrm{pts}$); (4) Return to the original lane ahead of the leader. (+$40\,\mathrm{pts}$)).
\end{itemize}
Across all scenarios that use graded scoring, common safety termination conditions are enforced. These include \textbf{red light violations}, \textbf{lane invasions}, and \textbf{collisions}. If any of these occur, the scenario is immediately terminated and the score accumulated up to that point is reported. This ensures that unsafe behaviors are penalized while preserving diagnostic resolution. Table~\ref{tab:driving_vln_scenarios} lists all driving VLN scenarios, their associated scoring schemes, and the thresholds applied. Scenario names below refer to the naming of scenarios in the code to be released upon publication.

\begin{table}[!t]
    \centering
    \scriptsize
    \resizebox{\linewidth}{!}{
    \begin{tabular}{c|l|c|cl}
        \toprule
        \textbf{ID} & \textbf{Scenario} & \textbf{Category} & \textbf{Binary} & \textbf{Graded; $\Theta$} \\
        \midrule
        D1.1 & HandbrakeLowbeam         & D1 & \cmark & -- \\
        D1.2 & SteeringSequence         & D1 & \cmark & -- \\
        D1.3 & Accelerate               & D1 & \cmark & -- \\
        D1.4 & MaintainSpeed            & D1 & \cmark & -- \\
        D1.5 & DistanceStop             & D1 & \cmark & -- \\
        D2.1 & Parking                  & D2 & -- & $\epsilon \leq 1.5\,\mathrm{m} \ \land \ v \leq 0.1\,\mathrm{m/s}$ \\
        D2.2 & ClearIntersection        & D2 & -- & $\epsilon \leq 5\,\mathrm{m} \ \land \ v \leq 0.1\,\mathrm{m/s} \land \text{laneID}$ \\
        D2.3 & Roundabout               & D2 & -- & $\epsilon \leq 5\,\mathrm{m} \ \land \ \text{laneID}$ \\
        D2.4 & TurnLeft                 & D2 & -- & $\epsilon \leq 5\,\mathrm{m} \ \land \ \text{laneID}$ \\
        D2.5 & LeaveHighway             & D2 & -- & $\epsilon \leq 5\,\mathrm{m} \ \land \ \text{laneID}$ \\
        D3.1 & EasySafeDrive            & D3 & -- & $d \geq 200\,\mathrm{m}$ \\
        D3.2 & FollowLane               & D3 & -- & $d \geq 210\,\mathrm{m} \ \land \ \text{laneID}$ \\
        D3.3 & FollowVehicle            & D3 & -- & $\epsilon \leq 15\,\mathrm{m} \ \land \ d \geq 100\,\mathrm{m}$ \\
        D3.4 & FollowVaryingVehicle     & D3 & -- & $\epsilon \leq 20\,\mathrm{m} \ \land \ d \geq 150\,\mathrm{m}$ \\
        D4.1 & SpeedLimit               & D4 & -- & $t \geq 10\,\mathrm{s} \ \land \ 80 \leq v \leq 90\,\mathrm{kph}$ \\
        D4.2 & Overtake                 & D4 & -- & see Section~\ref{appendix:vln:ad} \\
        \bottomrule
    \end{tabular}}
    \caption{\textbf{Autonomous driving VLN scenarios} with corresponding scoring schemes and thresholds. $\Theta$ denotes the $100$-points target condition if not scored binary.}
    \label{tab:driving_vln_scenarios}
\end{table}

\subsection{Aerial Navigation} %%%%%%%%%%%%%%%
\label{appendix:vln:an}

The aerial sub-benchmark follows the same scoring principles but adapts them to the UAV action space. It comprises three scenario categories of increasing instruction and task complexity, progressing from fundamental control understanding to full mission-level flight execution:

\begin{itemize}
    \item \textbf{(A1) Basic Control:}
    These tasks assess the UAV's fundamental control understanding through targeted single-axis maneuvers (e.g., \textit{``descend $3.2\,\mathrm{m}$"}). A1 scenarios are varied in both action direction and magnitude within the UAV's control space. Each task includes a single success condition, evaluated using binary $0/100$ scoring (cf. Eq.~\ref{eq:vlnbinary}).
    \item \textbf{(A2) Goal-Directed Navigation:}
    Goal-directed navigation tasks require the UAV to execute compound flight sequences that jointly control altitude, orientation, and translation (e.g., \textit{``ascend $12\,\mathrm{m}$, turn $70^\circ$ left, and continue straight for $10\,\mathrm{m}$ while maintaining altitude"}). Each sequence implicitly defines a spatial goal state in 3D airspace relative to the UAV's initial position $d_{\mathrm{init}}$. Progress is evaluated through a graded scoring scheme that computes the Euclidean distance between the UAV's final and target positions $d_{\mathrm{agt}}$, normalized with respect to $d_{\mathrm{init}}$, linearly distributing up to $50$ points across this range until the UAV reaches the $100$-point termination threshold (cf. Eq.~\ref{eq:vlngraded}). A perfect score of $100$ is assigned when the UAV terminates within $\theta_{\mathrm{agt}} \leq 1\,\mathrm{m}$ of the aerial target, reflecting the required spatial precision for controlled flight trajectories. To further encourage correct directional progress throughout the episode, the scoring is extended to continuously record the minimum achieved distance to the target, $d_{\mathrm{min}}$. The similar graded function is applied to this distance, producing a maximum progress score that rewards movement toward the correct aerial direction, even if the UAV overshoots the target. The final score is computed as a weighted combination of the terminal distance score and the maximum progress score, prioritizing successful goal completion (90\%) while maintaining sensitivity to directional accuracy (10\%).
    \item \textbf{(A3) Mission-level Execution:}
    These scenarios involve complex, high-level flight missions that demand coordinated spatial reasoning and temporal consistency, such as \textit{``LandInMiddleCircle"} or \textit{``HoverAboveCrocodile"}. Since mission completion either fails or succeeds, binary $0/100$ scoring is applied (cf. Eq.~\ref{eq:vlnbinary}). Partial progress is not rewarded, as incomplete execution typically violates the mission objective or safety constraints inherent to UAV operation. A scenario-specific target region defines successful completion, with thresholds detailed in Table\ref{tab:flying_vln_scenarios}. To ensure accurate vertical positioning, an additional altitude constraint $\delta_h$ is enforced, with $\lvert \delta_h \rvert \leq 1\,\mathrm{m}$. Exceeding this tolerance results in mission failure and a score of $0$.
\end{itemize}

\begin{table}[!t]
    \centering
    \scriptsize
    \resizebox{\linewidth}{!}{
    \begin{tabular}{c|l|c|cl}
        \toprule
        \textbf{ID} & \textbf{Scenario} & \textbf{Category} & \textbf{Binary} & \textbf{Graded; $\Theta$} \\
        \midrule
        A1.1-1.4    & SingleControlAction (x4) & A1 & \cmark & -- \\
        A2.1-2.12   & ControlSequence (x12)    & A2 & -- & $\epsilon \leq 1\,\mathrm{m}$ \\
        A3.1        & FlyToMerryGoAround       & A3 & \cmark & $\epsilon \leq 5\,\mathrm{m} \ \land \ \lvert \delta_h \rvert \leq 1\,\mathrm{m}$ \\
        A3.2        & LandInMiddleCircle       & A3 & \cmark & $\epsilon \leq 10\,\mathrm{m} \ \land \ \lvert \delta_h \rvert \leq 1\,\mathrm{m}$ \\
        A3.3        & LandOnRefereeShelter     & A3 & \cmark & $\epsilon \leq 3.5\,\mathrm{m} \ \land \ \lvert \delta_h \rvert \leq 1\,\mathrm{m}$ \\
        A3.4        & HoverAboveCrocodile      & A3 & \cmark & $\epsilon \leq 4\,\mathrm{m} \ \land \ \lvert \delta_h \rvert \leq 1\,\mathrm{m}$ \\
        A3.5        & LandOnTrashCan           & A3 & \cmark & $\epsilon \leq 1.5\,\mathrm{m} \ \land \ \lvert \delta_h \rvert \leq 1\,\mathrm{m}$ \\
        A3.6        & SearchTunnelEntrance     & A3 & \cmark & $\epsilon \leq 15\,\mathrm{m} \ \land \ \lvert \delta_h \rvert \leq 1\,\mathrm{m}$ \\
        \bottomrule
    \end{tabular}}
    \caption{\textbf{Aerial navigation VLN scenarios} with corresponding scoring schemes and thresholds. $\Theta$ denotes the $100$-points target condition.}
    \label{tab:flying_vln_scenarios}
\end{table}

\subsection{Robotic Manipulation} %%%%%%%%%%%%%%%
\label{appendix:vln:mp}

The manipulation sub-benchmark defines three scenario categories reflecting increasing complexity in language-guided robotic control. Table~\ref{tab:manipulation_vln_scenarios} lists all scenarios, their categories, and associated scoring schemes.

\begin{itemize}
    \item \textbf{(M1) Basic Control:} 
    These are simple, single-step tasks involving the actuation of a single joint or the gripper (e.g., \textit{open/close gripper}, \textit{move one joint}). Each M1 scenario is randomly varied within the robot's action space, but only a single action is required for success. Success is evaluated using a binary $0/100$ criterion (cf. Eq.~\ref{eq:vlnbinary}).
    \item \textbf{(M2) Goal-Directed Navigation:}
    These tasks involve multi-step, sequence-based navigation of the end-effector (e.g., \textit{``move $30\,\mathrm{cm}$ forward, then $30\,\mathrm{cm}$ right, then $30\,\mathrm{cm}$ down"}). Each sequence is randomly shuffled along the three orthogonal axes, so that while the distances remain fixed at $30\,\mathrm{cm}$ per step, the order of the directions and end target location varies. Graded scoring is applied: partial credit of up to $50$ points is awarded based on the reduction of the initial distance to the target ($d_{\mathrm{init}} = 51.9\,\mathrm{cm}$), and full $100$ points are granted only if the agent reaches within $\theta_{\mathrm{agt}} \leq 2.5\,\mathrm{cm}$ of the goal (cf. Eq.~\ref{eq:vlngraded}). This strict tolerance reflects the high precision required for manipulation, where even small deviations can prevent task success.
    \item \textbf{(M3) Dynamic Interaction:}
    Complex tasks such as \textit{BasketballInHoop}, \textit{PushButtons}, and \textit{OpenJar} demand precise, sequential coordination to achieve a defined goal. Since each task represents a single, well-defined mission (e.g., the ball either lands in the hoop or not), outcomes are evaluated using a binary $0/100$ score (cf. Eq.\ref{eq:vlnbinary}). Partial progress is not rewarded, as success depends on completing the full objective.    
\end{itemize}

\begin{table}[!t]
    \centering
    \scriptsize
    \resizebox{\linewidth}{!}{
    \begin{tabular}{c|l|c|cl}
        \toprule
        \textbf{ID} & \textbf{Scenario} & \textbf{Category} & \textbf{Binary} & \textbf{Graded; $\Theta$} \\
        \midrule
        M1.1-1.2    & SingleControlAction (x2) & M1 & \cmark & -- \\
        M2.1-2.12   & ControlSequence (x12)    & M2 & -- & $\epsilon \leq 2.5\,\mathrm{cm}$ \\
        M3.1        & BasketballInHoop         & M3 & \cmark & -- \\
        M3.2        & PutRubbishInBin          & M3 & \cmark & -- \\
        M3.3        & CloseLaptopLid           & M3 & \cmark & -- \\
        M3.4        & PushButtons              & M3 & \cmark & -- \\
        M3.5        & OpenJar                  & M3 & \cmark & -- \\
        M3.6        & ReachTarget              & M3 & \cmark & -- \\
        \bottomrule
    \end{tabular}}
    \caption{\textbf{Robotic manipulation VLN scenarios} with corresponding scoring schemes and thresholds. $\Theta$ denotes the $100$-points target condition.}
    \label{tab:manipulation_vln_scenarios}
\end{table}

%%%%%%%%%%%%%%%%%%%%%%%%%%%%%% AGENT MODELS %%%%%%%%%%%%%%%%%%%%%%%%%%%%%%
\section{Autonomous Agent Models} %%%%%%%%%%%%%%%
\label{appendix:autonomousagentschoice}

\subsection{Driving Agent Models} %%%%%%%%%%%%%%%

We select \textbf{Senna}~\cite{Jiang_2024_arXiv_Senna} as our driving baseline model due to its combination of open-source availability (OSA:~\cmark), and its support for interactive VQA, VLN, and spatial understanding. While models like DriveVLM~\cite{Tian_2024_arXiv_DriveVLM} and OpenDriveVLA~\cite{Zhou_2025_arXiv_OpenDriveVLA} offer advanced reasoning or multimodal fusion, they lack accessible code or inference pipelines (cf. Table~\ref{tab:drivingModels}). Senna's public implementation enables reproducible evaluation of interactive driving.

\textbf{Senna} is a composite vision-language driving model in which the core decision-making logic is realized by Senna-VLM. The backbone of Senna-VLM is based on Vicuna-7B-v1.5 following the LLaVA architecture paradigm, where a vision encoder is coupled to the language model through a \emph{driving-specific multimodal adapter}. This design enables Senna-VLM to generate structured natural language outputs that articulate driving intent and scene interpretation.
Senna operates on a fixed surround-view camera configuration of six images to maintain global spatial awareness. The model assumes this input dimensionality as part of its internal feature alignment; deviations from this requirement lead to inference failure. To accommodate varying real-world sensor setups of Embodied4C, we enforce a strict input normalization step: if fewer than six views are available, a placeholder (dummy) frame is inserted, and if more than six are present, the input is truncated. All experiments are conducted with Senna in 4-bit precision.

\begin{table}[!t]
    \centering
    \scriptsize
    \renewcommand{\arraystretch}{1.2}
    \resizebox{\linewidth}{!}{
    \begin{tabular}{r|cccc|cc|c}
        \toprule
        \textbf{Model}       
        & \textbf{SEM} & \textbf{SPA} & \textbf{TEM} & \textbf{PHY} & \textbf{INT} & \textbf{VLN} & \textbf{OSA} \\
        \midrule
        LanguageMPC~\cite{Sha_2023_arXiv_LanguageMPC}       & \xmark & \xmark & \xmark & \xmark & \xmark & \xmark & \xmark \\ % https://sites.google.com/view/llm-ad (no code only project page)
        DME-Driver~\cite{Han_2024_arXiv_DME-Driver}         & \xmark & \cmark & \xmark & \xmark & \xmark & \xmark & \xmark \\ % n.a.
        VLM-AD~\cite{Xu_2024_arXiv_VLM-AD}                  & \xmark & \xmark & \cmark & \xmark & \xmark & \xmark & \xmark \\ % n.a.
        GPT-Driver~\cite{Mao_2023_arXiv_GPT-Driver}         & \xmark & \xmark & \xmark & \xmark & \xmark & \xmark & \cmark \\ % https://github.com/PointsCoder/GPT-Driver
        OpenEMMA~\cite{Xing_2025_arXiv_OpenEMMA}            & \xmark & \xmark & \xmark & \xmark & \xmark & \xmark & \cmark \\ % https://github.com/taco-group/OpenEMMA
        LightEMMA~\cite{Qiao_2025_arXiv_LightEMMA}          & \xmark & \xmark & \xmark & \xmark & \xmark & \xmark & \cmark \\ % https://github.com/michigan-traffic-lab/LightEMMA
        EMMA~\cite{Hwang_2024_arXiv_EMMA}                   & \xmark & \xmark & \xmark & \xmark & \cmark & \xmark & \xmark \\ % n.a.
        Reason2Drive~\cite{Nie_2025_ECCV_Reason2Drive}      & \xmark & \xmark & \xmark & \xmark & \cmark & \xmark & \xmark \\ % https://github.com/fudan-zvg/Reason2Drive (only bench)
        DriveVLM~\cite{Tian_2024_arXiv_DriveVLM}            & \xmark & \xmark & \xmark & \xmark & \cmark & \xmark & \xmark \\ % https://tsinghua-mars-lab.github.io/DriveVLM/ (no model code)
        InternVL4Drive-v2~\cite{Li_2024_arXiv_InternVL}     & \xmark & \xmark & \cmark & \xmark & \cmark & \xmark & \xmark \\ % n.a.
        LaVida Drive~\cite{Jiao_2025_arXiv_LaVidaDrive}     & \xmark & \cmark & \cmark & \xmark & \cmark & \xmark & \xmark \\ % n.a.
        BEV-InMLLM~\cite{Ding_2024_CVPR_BEV-InMLLM}         & \xmark & \cmark & \cmark & \xmark & \cmark & \xmark & \xmark \\ % https://github.com/xmed-lab/NuInstruct (only NuInstruct dataset, no drive model)
        LiDAR-LLM~\cite{Yang_2023_arXiv_LiDAR-LLM}          & \xmark & \cmark & \xmark & \xmark & \cmark & \xmark & \xmark \\ % https://github.com/Yangsenqiao/LiDAR-LLM (only nu-Caption and nu-Grounding datasets, no model)
        OpenDriveVLA~\cite{Zhou_2025_arXiv_OpenDriveVLA}    & \cmark & \cmark & \xmark & \xmark & \cmark & \xmark & \xmark \\ % https://github.com/DriveVLA/OpenDriveVLA (missing inference code)
        OccLLaMA~\cite{Wei_2024_arXiv_OccLLaMA}             & \cmark & \cmark & \cmark & \xmark & \cmark & \xmark & \xmark \\ % n.a.
        DriverAgent~\cite{Jin_2024_arXiv_SurrealDriver}     & \xmark & \xmark & \xmark & \xmark & \xmark & \cmark & \xmark \\ % https://github.com/AIR-DISCOVER/Driving-Thinking-Dataset (only dataset)
        CarLLaVA~\cite{Renz_2024_arXiv_CarLLaVA}            & \xmark & \xmark & \xmark & \xmark & \xmark & \cmark & \xmark \\ % n.a.
        HE-Drive~\cite{Wang_2024_arXiv_HE-Drive}            & \xmark & \xmark & \xmark & \xmark & \cmark & \xmark & \cmark \\ % https://github.com/jmwang0117/HE-Drive
        RDA-Driver~\cite{Huang_2024_ECCV_RDA-Driver}        & \xmark & \xmark & \xmark & \xmark & \cmark & \xmark & \cmark \\ % https://github.com/zhijian11/RDA-Driver
        LLM-Driver~\cite{Chen_2024_ICRA_LLM-Driver}         & \xmark & \xmark & \xmark & \xmark & \cmark & \xmark & \cmark \\ % https://github.com/wayveai/Driving-with-LLMs
        DriveMM~\cite{Huang_2024_arXiv_DriveMM}             & \xmark & \xmark & \xmark & \xmark & \cmark & \xmark & \cmark \\ % https://github.com/zhijian11/DriveMM
        DriveLM-Agent~\cite{Sima_2025_ECCV_DriveLM}         & \xmark & \xmark & \xmark & \xmark & \cmark & \xmark & \cmark \\ % https://github.com/OpenDriveLab/DriveLM
        RAG-Driver~\cite{Yuan_2024_arXiv_RAG-Driver}        & \xmark & \xmark & \xmark & \xmark & \cmark & \xmark & \cmark \\ % https://github.com/YuanJianhao508/RAG-Driver
        BeVLM~\cite{Dong_2024_GitHub_BeVLM}                 & \xmark & \cmark & \xmark & \xmark & \cmark & \xmark & \cmark \\ % https://github.com/BEVLM/DrivingWithLanguage
        Dolphins~\cite{Yingzi_2025_ECCV_Dolphins}           & \cmark & \cmark & \xmark & \cmark & \cmark & \xmark & \cmark \\ % https://github.com/SaFoLab-WISC/Dolphins
        DriveGPT4~\cite{Xu_2024_IEEE_DriveGPT4}             & \xmark & \cmark & \cmark & \xmark & \cmark & \xmark & \cmark \\ % https://tonyxuqaq.github.io/projects/DriveGPT4/
        OmniDrive-Agent~\cite{Wang_2024_arXiv_OmniDrive}    & \xmark & \cmark & \cmark & \xmark & \cmark & \xmark & \cmark \\ % https://github.com/NVlabs/OmniDrive
        ELM~\cite{Zhou_2025_ECCV_ELM}                       & \xmark & \cmark & \cmark & \xmark & \cmark & \xmark & \cmark \\ % https://github.com/OpenDriveLab/ELM
        LMDrive~\cite{Shao_2024_CVPR_LangAuto_LMDrive}      & \xmark & \xmark & \xmark & \xmark & \xmark & \cmark & \cmark \\ % https://github.com/opendilab/LMDrive
        Agent-Driver~\cite{Mao_2024_arXiv_AgentDriver}      & \xmark & \xmark & \xmark & \xmark & \xmark & \cmark & \cmark \\ % https://github.com/USC-GVL/Agent-Driver
        DriveMLM~\cite{Wang_2023_arXiv_DriveMLM}            & \xmark & \xmark & \xmark & \xmark & \cmark & \cmark & \xmark \\ % https://github.com/OpenGVLab/DriveMLM (no model code)
        ORION~\cite{Fu_2025_arXiv_ORION}                    & \xmark & \xmark & \cmark & \xmark & \cmark & \cmark & \xmark \\ % https://github.com/xiaomi-mlab/Orion (no model code)
        Senna~\cite{Jiang_2024_arXiv_Senna}                 & \xmark & \cmark & \xmark & \xmark & \cmark & \cmark & \cmark \\ % https://github.com/hustvl/Senna
        SimLingo~\cite{Renz_2025_arXiv_SimLingo}            & \xmark & \cmark & \xmark & \xmark & \cmark & \cmark & \cmark \\ % https://github.com/RenzKa/simlingo
        AsyncDriver~\cite{Chen_2024_ECCV_AsyncDriver}       & \cmark & \xmark & \xmark & \xmark & \cmark & \cmark & \cmark \\ % https://github.com/memberRE/AsyncDriver
        \bottomrule
    \end{tabular}}
    \caption{\textbf{Overview of state-of-the-art driving agent models.} Columns ``\textbf{SEM}", ``\textbf{SPA}", ``\textbf{TEM}", and ``\textbf{PHY}" indicate the inclusion of semantic, spatial, temporal, and physical understanding. ``\textbf{INT}" and ``\textbf{VLN}" denote support for interaction through VQA and vision-language navigation, while ``\textbf{OSA}" marks open source availability (as of August 2025).}
    \label{tab:drivingModels}
\end{table}

\subsection{UAV Agent Models} %%%%%%%%%%%%%%%
We adopt \textbf{OpenFly-Agent}~\cite{Gao_2025_arXiv_OpenFly} as the UAV baseline. Among current models (cf. Table~\ref{tab:aerialModels}), it is one of the few that is fully open-source (\textbf{OSA:}~\cmark) and publicly released on Hugging Face, enabling transparent and reproducible evaluation. In contrast, most UAV VLN models either lack released checkpoints, rely on proprietary training pipelines, or provide perception-only components without actionable control policies.

\textbf{OpenFly-Agent} integrates a visual encoder with a LLaMA-derived language model and outputs a 6-dimensional action vector:  
\[
[\text{forward}, \text{turn\_left}, \text{turn\_right}, \text{move\_up}, \text{move\_down}, \text{stop}],
\]  
where each action corresponds to a fixed displacement of $3\,\mathrm{m}$ or a rotation step of $30^\circ$. For VLN tasks, we map these commands to the unified low-level control interface of each sub-benchmark, ensuring consistent actuation across agent setups. For VQA tasks, OpenFly's action head is bypassed to force textual responses, supported by an additional, minimal system instruction that enforces concise English output. Since the language model backbone was trained solely for action vector generation, it produces nonsensical text outputs, yielding zero scores (cf. Section~\ref{sec:experiments}). All experiments are conducted with Open-Fly Agent in 4-bit precision.

\begin{table}[!t]
    \centering
    \scriptsize
    \renewcommand{\arraystretch}{1.2}
    \resizebox{\linewidth}{!}{
    \begin{tabular}{r|cccc|cc|c}
        \toprule
        \textbf{Model}       
        & \textbf{SEM} & \textbf{SPA} & \textbf{TEM} & \textbf{PHY} & \textbf{INT} & \textbf{VLN} & \textbf{OSA} \\
        \midrule
        CognitiveDrone~\cite{Lykov_2025_arXiv_CognitiveDrone}   & \cmark & \xmark & \xmark & \xmark & \xmark & \cmark & \xmark \\ % https://github.com/SerValera/docker_CognitiveDrone_DataCollector // https://huggingface.co/datasets/ArtemLykov/CognitiveDrone_dataset(only dataset and data collector)
        LogisticsVLN~\cite{Zhang_2025_arXiv_LogisticsVLN}       & \cmark & \cmark & \xmark & \xmark & \xmark & \cmark & \xmark \\ % nothing
        VLFly~\cite{Zhang_2025_arXiv_VLFly}                     & \cmark & \cmark & \xmark & \xmark & \xmark & \cmark & \xmark \\ % https://github.com/zzzzzyh111/Vision-Language-Fly (no code yet)
        UAV-VLN~\cite{Saxena_2025_arXiv_UAV-VLN}                & \cmark & \cmark & \xmark & \xmark & \xmark & \cmark & \xmark \\ % nothing
        UAV-VLA~\cite{Sautenkov_2025_HRI_UAV-VLA}               & \cmark & \cmark & \xmark & \xmark & \xmark & \xmark & \cmark \\ % https://github.com/sautenich/uav-vla (but model cannot fly)
        LAG~\cite{Liu_2023_arXiv_AerialVLN}                     & \cmark & \cmark & \xmark & \xmark & \xmark & \cmark & \cmark \\ % https://github.com/AirVLN/AirVLN // download model from here: https://github.com/facebookresearch/habitat-lab/tree/v0.1.7/habitat_baselines/rl/ddppo
        OpenFly-Agent~\cite{Gao_2025_arXiv_OpenFly}             & \cmark & \cmark & \cmark & \xmark & \xmark & \cmark & \cmark \\ % https://github.com/SHAILAB-IPEC/OpenFly-Platform // Hugginface model: https://huggingface.co/IPEC-COMMUNITY/openfly-agent-7b
        \bottomrule
    \end{tabular}}
    \caption{\textbf{Overview of state-of-the-art UAV agent models.} Columns ``\textbf{SEM}", ``\textbf{SPA}", ``\textbf{TEM}", and ``\textbf{PHY}" indicate the inclusion of semantic, spatial, temporal, and physical understanding. ``\textbf{INT}" and ``\textbf{VLN}" denote support for interaction through VQA and vision-language navigation, while ``\textbf{OSA}" marks open source availability (as of September 2025).}
    \label{tab:aerialModels}
\end{table}

\subsection{Manipulation Agent Models} %%%%%%%%%%%%%%%
For manipulation tasks, we adopt two open-source VLA baselines: \textbf{OpenVLA}~\cite{Kim_2024_arXiv_OpenVLA} and \textbf{MolmoAct}~\cite{Lee_2025_arXiv_MolmoAct}, both available through public checkpoints (OSA:~\cmark) and thus suitable for reproducible, controlled comparison (cf.~Table~\ref{tab:manipulatorModels}). Closed-source systems, such as LA-RCS~\cite{Park_2025_arXiv_LARCS} and RT-2~\cite{Zitkovich_2023_PMLR_RT2}, are excluded due to deployment limitations.

\textbf{OpenVLA} is trained on large-scale robotic demonstration datasets and predicts continuous 7-DoF control commands for end-effector translation $(\Delta x, \Delta y, \Delta z)$, rotation $(\Delta \alpha, \Delta \beta, \Delta \gamma)$, and gripper state $g$. To ensure consistent execution across our sub-benchmark environments, we map these outputs into the standardized control interface and apply scaling adjustment. For VQA-style tasks, the action head is bypassed, and the language component is prompted directly to produce concise English answers using a minimal instruction. However, since the language backbone was trained solely for action prediction, direct text decoding produces nonsensical outputs, resulting in uniform zero scores. All experiments are conducted with OpenVLA in 4-bit precision.
\textbf{MolmoAct} similarly provides a 7-DoF action head but supports explicit switching between action prediction and text generation. We use the \texttt{MolmoAct-7B-D-0812} release and provide an additional alignment prompt to standardize its action dimensions to corresponding sub-benchmark interfaces. For VQA tasks, we directly decode language tokens, while VLN tasks use the native action outputs.

This setup allows consistent cross-model comparison of scenario understanding among the four capability dimensions and action execution.

\begin{table}[!t]
    \centering
    \scriptsize
    \renewcommand{\arraystretch}{1.2}
    \resizebox{\linewidth}{!}{
    \begin{tabular}{r|cccc|cc|c}
        \toprule
        \textbf{Model}       
        & \textbf{SEM} & \textbf{SPA} & \textbf{TEM} & \textbf{PHY} & \textbf{INT} & \textbf{VLN} & \textbf{OSA} \\
        \midrule
        MALMM~\cite{Singh_2025_arXiv_MALMM}         & \cmark & \cmark & \xmark & \xmark & \xmark & \cmark & \xmark \\ % https://github.com/malmm1/MALMM (empty)
        LA-RCS~\cite{Park_2025_arXiv_LARCS}         & \cmark & \cmark & \cmark & \xmark & \cmark & \cmark & \xmark \\ % https://la-rcs.github.io/ (only project page, no code)
        RT-2~\cite{Zitkovich_2023_PMLR_RT2}         & \cmark & \cmark & \xmark & \cmark & \cmark & \cmark & \xmark \\ % https://robotics-transformer2.github.io/ (only project page, no code)
        CLIPort~\cite{Shridhar_2022_PMLR_CLIPort}   & \cmark & \cmark & \xmark & \xmark & \xmark & \cmark & \cmark \\ % https://github.com/cliport/cliport (needs to be trained)
        VoxPoser~\cite{Huang_2023_arXiv_VoxPoser}   & \cmark & \cmark & \xmark & \xmark & \xmark & \cmark & \cmark \\ % https://github.com/huangwl18/VoxPoser (extra für RLBench!!!)
        RT-1~\cite{Brohan_2023_arXiv_RT1}           & \cmark & \cmark & \xmark & \xmark & \xmark & \cmark & \cmark \\ % https://github.com/google-research/robotics_transformer
        SpatialVLA~\cite{Qu_2025_arXiv_SpatialVLA}  & \cmark & \cmark & \xmark & \xmark & \xmark & \cmark & \cmark \\ % https://github.com/SpatialVLA/SpatialVLA (für SimplerEnv gut)
        OpenVLA~\cite{Kim_2024_arXiv_OpenVLA}       & \cmark & \cmark & \xmark & \xmark & \xmark & \cmark & \cmark \\ % https://github.com/openvla/openvla
        MolmoAct~\cite{Lee_2025_arXiv_MolmoAct}     & \cmark & \cmark & \xmark & \xmark & \xmark & \cmark & \cmark \\ % https://huggingface.co/allenai/MolmoAct-7B-D-0812 (Huggingface modell!!!)
        \bottomrule
    \end{tabular}}
    \caption{\textbf{Overview of state-of-the-art manipulation agent models.} Columns ``\textbf{SEM}", ``\textbf{SPA}", ``\textbf{TEM}", and ``\textbf{PHY}" indicate the inclusion of semantic, spatial, temporal, and physical understanding. ``\textbf{INT}" and ``\textbf{VLN}" denote support for interaction through VQA and vision-language navigation, while ``\textbf{OSA}" marks open source availability (as of September 2025).}
    \label{tab:manipulatorModels}
\end{table}

\begin{figure*}[!t]
    \centering
    \includegraphics[width=\linewidth]{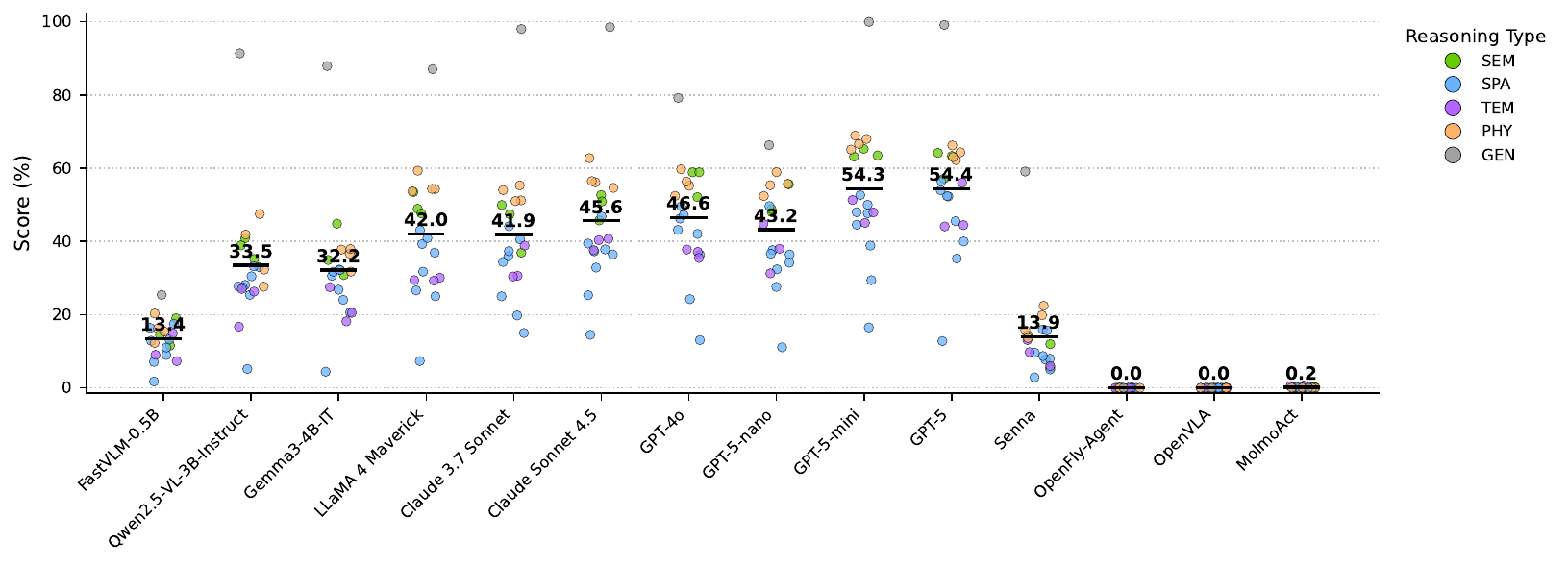}
    \caption{\textbf{Model robustness across Embodied4C sub-capabilities.} Each dot represents a model's mean score on one sub-capability (19 total), jittered horizontally for clarity; the black bar marks the model's overall VQA mean. Color encodes high-level reasoning type: semantic, spatial, temporal, physical, and generalization.}
    \label{fig:variances}
\end{figure*}

\begin{figure}[!t]
    \centering
    \includegraphics[width=\linewidth]{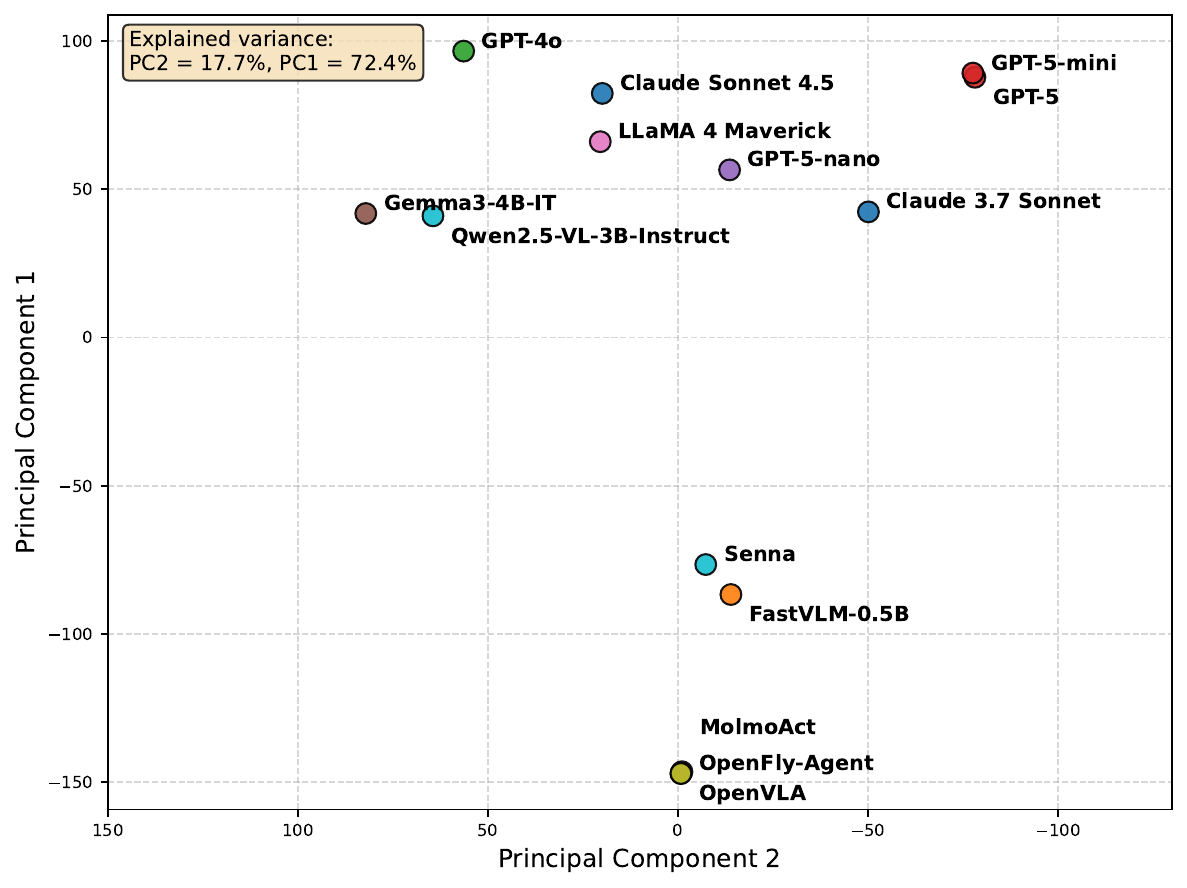}
    \caption{\textbf{Principle component analysis of Embodied4C model capability profiles.} PC1 (72.4\%) reflects overall embodied competence, separating low-level visuomotor models from high-capability multimodal reasoners. PC2 (17.7\%) distinguishes reasoning strategy, ranging from language-driven planning to direct perception-to-action control.}
    \label{fig:pca}
\end{figure}

%%%%%%%%%%%%%%%%%%%%%%%%%%%%%% VISION LANGUAGE NAVIGATION %%%%%%%%%%%%%%%%%%%%%%%%%%%%%%
\section{Additional Results}
\label{appendix:results}

\subsection{Sub-Capability Analysis}
\label{appendix:results:heatmap}

The heatmap summarizes performance across the 18 sub-capabilities of Embodied4C and the generalization category (GEN). Horizontally (across models), a clear scaling trend emerges: large frontier models (GPT-5~\cite{OpenAI_2025_GPT5}, Claude~Sonnet~4.5~\cite{Anthropic_2025_Claude4}, LLaMA~4~Maverick~\cite{Meta_2024_Blog_Llama4}) achieve substantially higher average performance, while smaller models ($\leq$7B) show consistently lower scores. Domain-specialized embodied agents fail significantly across all sub-capabilities, as training distributions and bypassing their action heads yields bad or no coherent language output.

To further characterize cross-capability stability (cf. Figure~\ref{fig:variances}), we analyze the coefficient of variation (CV), defined as:
\begin{equation}
   \mathrm{CV} = \frac{\sigma}{\mu}, \quad \mathrm{CV}\% = 100 \cdot \frac{\sigma}{\mu}, 
\end{equation}
where $\mu$ is the mean performance over sub-capabilities and $\sigma$ is the standard deviation. CV provides a scale-normalized measure of dispersion, enabling fair comparison between models with different absolute score ranges. Smaller models (e.g., Qwen2.5-VL-3B-Instruct~\cite{Qwen2.5-VL}, Gemma3-4B-IT~\cite{Google_2025_Gemma3}) exhibit high variability (CV$\approx$50\%), indicating inconsistent reasoning capabilities across the capability spectrum. In contrast, higher-capability frontier models demonstrate more stable behavior: GPT-5 achieves $\mu=54.4$, $\sigma=16.9$, CV$=31.1\%$, and GPT-4o~\cite{OpenAI_2024_arXiv_GPT-4o} similarly maintains lower dispersion (CV$=31.3\%$). This trend suggests that scaling not only improves average capability but also reduces cross-capability unevenness.

Vertically (across sub-capabilities), two clusters emerge. \textit{Semantic} and \textit{Physical} reasoning are comparatively stronger (up to $68.9\%$), particularly for model dynamics (MD) and environmental effects (E/E). Conversely, \textit{Spatial} and \textit{Temporal} reasoning remain major failure modes. Distance estimation (DT) is the weakest dimension overall ($\mu = 7.46$), followed by orientation (OR) and counting (CT); within \textit{Temporal}, short-term memory (S/M) outperforms movement-time (M/T) and long-term memory (L/M), but all remain low.

Finally, generalization (GEN) is near-saturated for the strongest models, reflecting robustness to domain-far linguistic variation. Notable deviations (e.g., GPT-4o scoring lower despite strong average performance) suggest sensitivity to distractors or prompt instability rather than inability to generalize. Table~\ref{tab:appendix:statisticaleval} provides additional statistical features across all tested models.

\begin{table*}[!t]
    \centering
    \footnotesize
    \begin{tabular}{l|cccccc|c}
        \toprule
        \textbf{Model} & \textbf{Mean} & \textbf{Std} & \textbf{Min} & \textbf{Max} & \textbf{Range} & \textbf{CV\%} & \textbf{$\#>0$} \\
        \midrule
        FastVLM-0.5B~\cite{FastVLM_2025_Apple}                  & 13.4 &  5.4 &  1.7 & 25.4 & 23.7 & 40.4\% & 19 \\
        Qwen2.5-VL-3B-Instruct~\cite{Qwen2.5-VL}                & 33.5 & 16.8 &  5.2 & 91.4 & 86.2 & 50.2\% & 19 \\
        Gemma3-4B-IT~\cite{Google_2025_Gemma3}                  & 32.2 & 16.2 &  4.4 & 87.9 & 83.5 & 50.3\% & 19 \\
        LLaMA 4 Maverick~\cite{Meta_2024_Blog_Llama4}           & 42.0 & 17.2 &  7.3 & 87.1 & 79.7 & 40.9\% & 19 \\
        Claude 3.7 Sonnet~\cite{Anthropic_2024_Website_Claude3} & 41.9 & 17.7 & 15.0 & 98.0 & 83.0 & 42.3\% & 19 \\
        Claude Sonnet 4.5~\cite{Anthropic_2025_Claude4}         & 45.6 & 17.3 & 14.5 & 98.5 & 84.0 & 37.9\% & 19 \\
        GPT-4o~\cite{OpenAI_2024_arXiv_GPT-4o}                  & 46.6 & 14.6 & 13.0 & 79.2 & 66.1 & 31.3\% & 19 \\
        GPT-5-nano~\cite{OpenAI_2025_GPT5}                      & 43.2 & 13.3 & 11.1 & 66.3 & 55.2 & 30.8\% & 19 \\
        GPT-5-mini~\cite{OpenAI_2025_GPT5}                      & 54.3 & 17.7 & 16.5 &100.0 & 83.5 & 32.6\% & 19 \\
        GPT-5~\cite{OpenAI_2025_GPT5}                           & 54.4 & 16.9 & 12.8 & 99.1 & 86.4 & 31.1\% & 19 \\
        \midrule
        Senna~\cite{Jiang_2024_arXiv_Senna}                     & 13.9 & 12.1 & 2.9 & 59.1 & 56.2 & 86.9\%  & 19 \\
        OpenFly-Agent~\cite{Gao_2025_arXiv_OpenFly}             & 0.0 &  0.0 &  0.0 &  0.0 &  0.0 & --     & 0 \\
        OpenVLA~\cite{Kim_2024_arXiv_OpenVLA}                   & 0.0 &  0.0 &  0.0 &  0.0 &  0.0 & --     & 0 \\
        MolmoAct~\cite{Lee_2025_arXiv_MolmoAct}                 & 0.2 &  0.2 &  0.0 &  0.6 &  0.6 & 115.8\% & 17 \\
        \bottomrule
    \end{tabular}
    \caption{\textbf{Per-model VQA performance across 19 sub-capabilities in Embodied4C.} Reported statistics (mean, standard deviation, min, max, range, coefficient of variation, and count of non-zero scores) summarize accuracy scores ($0$–$100\%$) over fine-grained scenario understanding capabilities and tasks spanning autonomous driving, UAV navigation, and robotic manipulation.}
    \label{tab:appendix:statisticaleval}
\end{table*}

\subsection{Principal Component Analysis}
\label{appendix:results:pca}

To understand the latent structure underlying model behavior in Embodied4C, we perform a PCA over model-level capability vectors spanning semantic, spatial, temporal, and physical reasoning across all embodied domains. The resulting projection identifies two orthogonal dimensions---PC1 and PC2---that together explain \textbf{90.1\%} of the total variance (\textbf{72.4\%} and \textbf{17.7\%}, respectively). These components reveal not only performance magnitude but also distinct strategies through which embodied reasoning competence emerges.

\paragraph{PC1: Overall Benchmarked Performance.}
The first principal component reflects a near-linear correlation with the overall \textbf{E4C-S} score and the averaged domain scores (DS, AS, MS). Models with high PC1 values---most notably GPT-5-mini, GPT-5, and Claude~Sonnet~4.5---exhibit the strongest integrated performance across all embodied domains, achieving the top E4C-S values of \textbf{39.59}, \textbf{36.00}, and \textbf{31.78}, respectively. In contrast, models such as Senna, OpenVLA, and MolmoAct cluster at the negative end of PC1 with E4C-S scores below \textbf{8.5}, indicating poor general reasoning and limited cross-modal transfer. PC1 thus quantifies \textit{global embodied competence}, representing the dominant axis of model performance variation across the benchmark.

\paragraph{PC2: Generality and Robustness vs. Core VQA Capability.}
The second component captures an orthogonal dimension that differentiates models by their \emph{strategy of achieving high performance}. It contrasts robustness and generalization strength against specialization in core VQA sub-capabilities. Negative PC2 values correspond to models exhibiting \textbf{high generality and robustness}---notably GPT-5 and GPT-5-mini, both scoring near-perfect on the Domain-far QA metric ($\approx 100\%$). Their performance reflects stable reasoning across unfamiliar scenarios and minimal overfitting. Positive PC2 values are associated with models whose strength lies in \textbf{core visuolinguistic reasoning}, such as GPT-4o, Claude~Sonnet~4.5, Gemma3-4B-IT, and Qwen2.5-VL, which achieve high spatial and temporal understanding but lower generalization.  
Importantly, this axis must be interpreted \textit{relative to overall capability (PC1)}: models with low E4C-S, such as Senna, may appear in the positive PC2 range due to unbalanced or noisy sub-capability patterns rather than genuine VQA specialization. PC2 therefore encodes a \textit{trade-off axis between generality and VQA specialization}, meaningful primarily when considered in conjunction with global competence along PC1.

\paragraph{Interpretation.}
The two principal components together delineate a coherent performance manifold across embodied foundation models.
\begin{itemize}
    \item The \textbf{upper region (high PC1)} corresponds to globally strong performers with well-integrated multimodal reasoning. 
    \item The \textbf{left side (low PC2)} identifies the most general and robust models (relative to PC1), resilient to domain shifts and linguistic perturbations.
    \item The \textbf{right side (high PC2)} captures models whose strength lies in precise, domain-specific VQA competence.  
    \item Finally, the \textbf{bottom region (low PC1)} groups weakly capable VLA models dominated by sensorimotor coupling without transferable reasoning.
\end{itemize}

In summary, PC1 measures the \textit{depth} of embodied reasoning, while PC2 measures its \textit{robustness–specialization trade-off}. The top-right quadrant, occupied by GPT-5 and GPT-5-mini, marks the region of \textbf{maximal embodied general intelligence}: high overall competence, balanced multimodal understanding, and robust generalization across domains.

\end{document}